\documentclass[10pt,journal,compsoc]{IEEEtran}

\usepackage{soul}
\usepackage{url}
\usepackage[hidelinks]{hyperref}
\usepackage[utf8]{inputenc}
\usepackage[small]{caption}
\usepackage{graphicx}
\usepackage{amsmath}
\usepackage{amsthm}
\usepackage{booktabs}
\usepackage{algorithm}
\usepackage{algorithmic}
\usepackage{comment}
\usepackage{listings}
\usepackage{xcolor}
\usepackage{tabularx}
\usepackage{booktabs}
\usepackage{multirow}
\usepackage{textcomp}
\usepackage{breqn}  
\usepackage{hyperref}
\usepackage{dblfloatfix} 
\usepackage{nicematrix}
\usepackage{amssymb}
\usepackage{bbm}
\usepackage{bm}
\usepackage{subcaption}

\usepackage{caption}

\begin{document}

\markboth{This paper is accepted in the \textbf{IEEE Transactions on Pattern Analysis and Machine Intelligence (TPAMI)}}%
	    {This paper is accepted in the \textbf{IEEE Transactions on Pattern Analysis and Machine Intelligence (TPAMI)}}

\title{Self-supervised Contrastive Representation Learning for Semi-supervised Time-Series Classification}

\author{Emadeldeen Eldele, Mohamed Ragab, Zhenghua Chen, Min Wu, Chee-Keong Kwoh, Xiaoli Li and Cuntai Guan~\IEEEmembership{Fellow,~IEEE}
\IEEEcompsocitemizethanks{
    \IEEEcompsocthanksitem{Emadeldeen Eldele is with the School of Computer Science and Engineering, Nanyang Technological University, Singapore and the Centre for Frontier AI Research, Agency for Science, Technology and Research, Singapore (E-mail: emad0002@ntu.edu.sg).}
    \IEEEcompsocthanksitem{Mohamed Ragab and Zhenghua Chen are with the Institute for Infocomm Research, Agency for Science, Technology and Research, Singapore, Centre for Frontier AI Research, Agency for Science, Technology and Research, Singapore (E-mail: \{mohamedr002, chen0832\}@e.ntu.edu.sg).}
    \IEEEcompsocthanksitem{Min Wu is with the Institute for Infocomm Research, Agency for Science, Technology and Research, Singapore (E-mail: wumin@i2r.a-star.edu.sg).}
    \IEEEcompsocthanksitem{Chee-Keong Kwoh and Cuntai Guan are with the School of Computer Science and Engineering, Nanyang Technological University, Singapore (E-mails: \{asckkwoh, ctguan\}@ntu.edu.sg).}
    \IEEEcompsocthanksitem{Xiaoli Li is with the Institute for Infocomm Research, Agency for Science, Technology and Research, Singapore, Centre for Frontier AI Research, Agency for Science, Technology and Research, Singapore, and also with the School of Computer Science and Engineering at Nanyang Technological University, Singapore (E-mail: xlli@i2r.a-star.edu.sg).}
    \IEEEcompsocthanksitem{The first author is supported by A*STAR SINGA Scholarship. Min Wu is the corresponding author.}
    \IEEEcompsocthanksitem{This research is supported by the Agency for Science, Technology and Research (A*STAR) under its AME Programmatic Funds (Grant No. A20H6b0151) and Career Development Award (Grant No. C210112046). It is also supported by the MOE Academic Research (Grant No: MOE2019-T2-2-175).}}
}

\IEEEtitleabstractindextext{

\begin{abstract}
Learning time-series representations when only unlabeled data or few labeled samples are available can be a challenging task. Recently, contrastive self-supervised learning has shown great improvement in extracting useful representations from unlabeled data via contrasting different augmented views of data. In this work, we propose a novel \textbf{T}ime-\textbf{S}eries representation learning framework via \textbf{T}emporal and \textbf{C}ontextual \textbf{C}ontrasting (\textbf{TS-TCC}) that learns representations from unlabeled data with contrastive learning. Specifically, we propose time-series-specific weak and strong augmentations and use their views to learn robust temporal relations in the proposed temporal contrasting module, besides learning discriminative representations by our proposed contextual contrasting module. Additionally, we conduct a systematic study of time-series data augmentation selection, which is a key part of contrastive learning. We also extend TS-TCC to the semi-supervised learning settings and propose a \textbf{C}lass-\textbf{A}ware TS-\textbf{TCC} (\textbf{CA-TCC}) that benefits from the available few labeled data to further improve representations learned by TS-TCC. Specifically, we leverage the robust pseudo labels produced by TS-TCC to realize a class-aware contrastive loss. Extensive experiments show that the linear evaluation of the features learned by our proposed framework performs comparably with the fully supervised training. Additionally, our framework shows high efficiency in few labeled data and transfer learning scenarios. The code is publicly available at \url{https://github.com/emadeldeen24/CA-TCC}.

\end{abstract}

}

\maketitle

\IEEEdisplaynontitleabstractindextext

\IEEEpeerreviewmaketitle

\ifCLASSOPTIONcompsoc
\IEEEraisesectionheading{\section{Introduction}\label{sec:introduction}}
\else
\section{Introduction}
\label{sec:introduction}
\fi

\IEEEPARstart{T}ime-series data are being incrementally collected on a daily basis from IoT, wearable devices, and machines sensors for various applications in healthcare, manufacturing, etc. \cite{gharehbaghi2017deep}.
However, unlike images, time-series data generally do not contain human recognizable patterns to differentiate different classes and thus easily assign class labels to data. Consequently, different time-series applications require trained specialists to perform the challenging data annotation/labeling task.
Therefore, little time-series data have been labeled in real-world applications compared to the collected data \cite{ching2018opportunities}.
With the advance of deep learning techniques, it becomes essential to train deep learning-based models with a massive amount of labeled data to learn useful representations and avoid overfitting. However, applying them to time-series data becomes challenging with the aforementioned labeling limitations.

Self-supervised learning has gained much attention recently for exploiting unlabeled data to learn powerful representations for deep learning models. Compared with models trained on fully labeled data (i.e., supervised models), self-supervised pretrained models can extract effective representations and achieve comparable performance when fine-tuned with a small percentage of the labels \cite{puzzle,simclr_paper}. The recent rebirth of self-supervised learning relied on manually-designed pretext tasks to learn representations about data. For example, Noroozi et al. split the image into smaller patches, shuffled them, and trained the model to reorder these patches to form the original image \cite{puzzle}. Some other pretext tasks assigned pseudo labels to different variations of input samples. For instance, Gidaris et al. applied several rotations to the original image and assigned a label to each angle \cite{gidaris_unsupervised}. The model was then trained to predict the rotation angle of the image as a classification task. 

Contrastive learning has recently shown its strong ability over pretext tasks to learn representations from unlabeled data. The strength point of contrastive learning is its ability to learn invariant representations by contrasting different views of the input sample, which are generated using augmentation techniques~\cite{simclr_paper,hjelm2018learning,He_2020_CVPR}. However, these image-based contrastive learning methods may not be able to work well on time-series data for the following reasons. First, unlike images, where features are mostly spatial, the time-series data are mainly characterized by their temporal dependencies~\cite{NEURIPS2019_53c6de78}. Therefore, applying the aforementioned techniques directly to time-series data may not efficiently address the temporal features of data. Second, some augmentation techniques used for images such as color distortion, generally cannot fit well with time-series data. 
So far, few works on contrastive learning have been proposed for time-series data. For example,~\cite{pmlr-v136-mohsenvand20a,cheng2020subject} developed contrastive learning methods for bio-signals. However, these two methods are proposed for specific clinical applications and may not generalize to other time-series data.

In this paper, we propose a novel framework that incorporates contrastive learning into self- and semi-supervised learning.
Specifically, we propose a \textbf{T}ime-\textbf{S}eries representation learning framework via \textbf{T}emporal and \textbf{C}ontextual \textbf{C}ontrasting (TS-TCC) that is trained on totally unlabeled datasets. Our TS-TCC employs two contrastive learning and augmentation techniques to handle the temporal dependencies of time-series data. We propose simple yet efficient data augmentations that can fit any time-series data to create two different, but correlated views of the input samples. These views are then used by the two innovative contrastive learning modules. In the first module, we propose a novel temporal contrasting module to learn \textit{robust} representations by designing a tough cross-view prediction task. Specifically, for a certain timestep, it utilizes the past latent features of one augmentation to predict the future of the other augmentation. This novel operation will force the model to learn robust representation by a harder prediction task against any perturbations introduced by different timesteps and augmentations. In the second module, we propose contextual contrasting to further learn \textit{discriminative} representations upon the robust representations learned by the temporal contrasting module. In this contextual contrasting module, we aim to maximize the similarity among different contexts of the same sample while minimizing similarity among contexts of different samples. The pretrained model learns powerful representations about the time-series data regardless of downstream tasks.

We also extend TS-TCC to the semi-supervised settings, where few labeled samples are available, and propose the \textbf{C}lass-\textbf{A}ware TS-TCC (\textbf{CA-TCC}). Specifically, we fine-tune the TS-TCC encoder with the few labeled samples and deploy it to generate pseudo labels for the whole unlabeled data. Next, we utilize the class information in the pseudo labels in the supervised contextual contrasting between samples to maximize the similarity between samples having the same class label, while minimizing the similarity between samples from different classes.

This journal paper is an extended version of our previous work \cite{eldele_ts_tcc}. The main adjustments are listed as follows: 
1) We extend our TS-TCC model to fit the semi-supervised setting by taking advantage of the available few labeled data to further improve the learned representations. 2) We study the time-series data augmentations, which are considered as a key part of contrastive learning, in more details. We also provide a systematic methodology for selecting the best augmentations for time-series data. 3) We conduct additional experiments to assess our proposed CA-TCC, demonstrating the effectiveness of its learned representations in semi-supervised settings. In summary, the main contributions of this work are as follows.


\begin{itemize}
    \item A novel contrastive learning framework for time-series representation learning with two variants is developed. The first (TS-TCC) is proposed for learning from unlabeled data, and the second (CA-TCC) is proposed for semi-supervised learning. 
    
    \item We provide simple yet efficient augmentations that are designed for time-series data in the contrastive learning framework, and provide extensive discussion about their selection. 
    
    \item We propose a novel temporal contrasting module to learn robust representations from time-series data by designing a tough cross-view prediction task. In addition, we propose a (supervised) contextual contrasting module to further learn discriminative representations upon the robust representations.
    
    \item We perform extensive experiments on the two proposed variants of our framework using ten real-world datasets. Experimental results show that both variants are capable of learning effective representations for different scenarios.

\end{itemize}

\section{Related Works}

\subsection{Self-supervised Learning}
Self-supervised learning is gaining more attention recently, as it deals with the emerging problem of learning from unlabeled data \cite{qi2020small,jing2020self,jaiswal2020a}.
The recent advances in self-supervised learning started with applying pretext tasks on images to learn useful representations.
Many pretext tasks, that vary in nature, have been proposed according to the target applications. For image-related applications, Zhang et al. trained their model to color the input grayscale image \cite{zhang2016colorful}. Misra et al. proposed learning invariant representations based on pretext tasks by encouraging the representation of both the pretext task and the original image to be similar \cite{misra2020self}. Zhai et al. developed pretext tasks based on self-supervised training to empower the representations learned in a semi-supervised setting \cite{Zhai_2019_ICCV}.

For video-related tasks, Srivastava et al. proposed different tasks such as input sequence reconstruction, or future sequence prediction to learn representations of video sequences~\cite{srivastava2015unsupervised}.
In addition, Wei et al. proposed training the model to check if the order of the input frame sequences is correct or not \cite{wei2018learning}.
Another direction is to use ranking as a proxy task to solve regression problems~\cite{liu2019exploiting}, in which the authors developed a backpropagation technique for Siamese networks to prevent redundant computations.
Although using pretext tasks can improve representation learning, they are found to limit the generality of the learned representations. For example, classifying the different rotation angles of an image may deviate the model from learning features about the color or orientation of objects~\cite{oord2018representation}.

On the other hand, contrastive methods intend to learn invariant representations from different augmented views of data. The different methods vary between each other in their ways of choosing negative samples against positive samples during training. SimCLR~\cite{simclr_paper} considered the augmented views of the original image as the positives, while all the other views of different images within the batch are treated as negatives. This technique benefits from larger batch sizes to accumulate more negative samples. Another approach was to accumulate a large number of the negative samples in a memory bank as proposed in MoCo~\cite{He_2020_CVPR}. The embeddings of these samples are updated with the most recent ones at regular intervals.
The next direction was to use contrastive learning without the need for negative samples. For example, BYOL~\cite{grill2020bootstrap} learned representations by bootstrapping representations from two neural networks
that interact and learn from each other. SimSiam~\cite{simsiam} supported the idea of neglecting the negative samples and relied only on a Siamese network and stop-gradient operation to achieve state-of-the-art performance.

Some methods deployed contrastive learning in semi-supervised settings. For example, FixMatch~\cite{sohn2020fixmatch} used weak augmented views to produce pseudo labels for the input image, and if the pseudo label exceeds a confidence threshold, they use it to penalize the prediction of the strong augmented view of the same input image. While all these approaches have successfully improved representation learning for visual data, they may not work well on time-series data that have different properties, such as temporal dependency.

\subsection{Self-supervised Learning for Time-Series}
Self-supervised representation learning for time-series is becoming more popular recently.
Some approaches employed pretext tasks for time-series data. 
For example, Saeed et al. designed a binary classification pretext task for human activity recognition by applying several transformations on the data and trained the model to classify between the original and the transformed versions \cite{SSL_har}.
Similarly, Sarkar et al. proposed SSL-ECG, in which ECG representations are learned by applying six transformations to the dataset as pretext tasks, and assigned pseudo labels according to the transformation type \cite{ecg_emotion_rec}. The model is expected to learn useful representations about the data by classifying these transformations.
The same approach was used by Saeed et al. as they designed eight auxiliary tasks and learning representations from multi-sensor human activity data \cite{saeed2020sense}.
Additionally, Aggarwal et al. learned subject-invariant representations by modeling local and global activity patterns \cite{aggarwal2019adversarial}. 

Inspired by the success of contrastive learning in visual applications, few works have recently leveraged contrastive learning for time-series data. 
For example, CPC \cite{oord2018representation} learned representations by predicting the future in the latent space and showed great advances in various speech recognition tasks.
Banville et al. studied three self-supervised tasks with two pretext tasks, i.e., relative positioning, temporal shuffling, and one contrastive task that uses CPC to learn representations about clinical EEG data  \cite{banville2020uncovering}. They found that representations learned by CPC perform the best. This indicates that contrastive learning techniques generally perform better than pretext tasks. 
Mohsenvand et al. designed EEG-related augmentations and extended the SimCLR model \cite{simclr_paper} to different EEG clinical data \cite{pmlr-v136-mohsenvand20a}. 
TS2Vec~\cite{yue2022ts2vec} proposed two contrastive losses, where the first pulls timesteps from the augmented views closer and pulls the different timesteps from the same time series away, while the second is the instance-wise contrasting.

Some methods also included the frequency domain characteristics to learn time-series representations. For example, Bilinear Temporal-Spectral Fusion (BTSF)~\cite{icml2022iterative} applied an iterative bilinear fusion between feature embeddings of both time and frequency representations of time series. Similarly, both TF-C~\cite{zhang2022_tfConsistency} and STFNets~\cite{STFNets} learned representations by pushing the time domain and frequency domain representations of the same sample closer to each other, while pushing them apart from representations of other signals. 

Existing approaches used either temporal or global features. Differently, we address both types of features in our cross-view temporal and contextual contrasting modules. These modules rely on different views for input data that we provide via time-series-specific augmentations.

\subsection{Semi-supervised Learning for Time-Series}
In semi-supervised learning, a model is trained on a dataset that is partially labeled, i.e., some samples in the dataset have been labeled, where it may be expensive or time-consuming to label all of the examples in the dataset. There are different approaches for time-series semi-supervised learning.

The first approach is to apply generic regularizations, e.g., ensembling and reconstruction.
For example, TEBLSTM~\cite{teblstm} applies temporal ensembling based on LSTM output features. It ensembles two losses, i.e., the supervised loss on the labeled data, and an unsupervised loss based on the distance of the current prediction and the previous ensemble output.
Also, REG-GAN~\cite{gan_regression} uses Generative Adversarial Networks (GAN) to learn representations by regenerating the signals of the whole data and applying supervised regression loss on the labeled part.

The second approach is self-training, where the unlabeled samples are assigned pseudo labels. For example, SUCCESS~\cite{success_semi} clustered the data, then assigned pseudo labels based on the Dynamic Time Warping (DTW) distance.
In addition, SelfMatch~\cite{selfMatch} deployed Gaussian noised signals to guide the prediction of the augmented view of the same signal. It also applied self-distillation by guiding the lower-level blocks of the feature extractor with the knowledge obtained in the output layer.
Last, SemiTime~\cite{semiTime} split the signal into two parts, i.e., past and future. Next, it applied contrastive loss to push the past of the signal toward its future, while pulling it away from the future parts of other signals.

\section{Methods}

\begin{figure}
\centering
\includegraphics[width=\columnwidth]{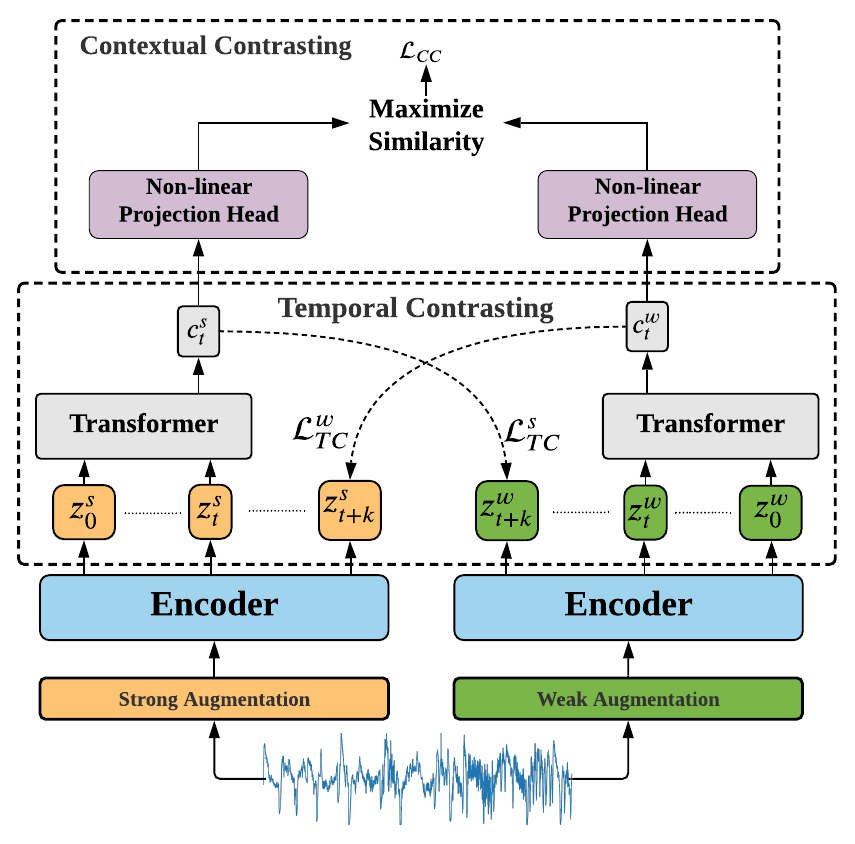}
\caption{The overall architecture of the proposed TS-TCC. The Temporal Contrasting module learns robust temporal features through a tough cross-view prediction task. The Contextual Contrasting module learns discriminative features by maximizing the similarity between the contexts of the same sample while minimizing its similarity with the other samples within the mini-batch.}
\label{Fig:overall}
\end{figure}

This section describes the components of the proposed framework in details. Starting with TS-TCC, we first generate two different yet correlated views of the input data based on strong and weak augmentations. Then, a temporal contrasting module is proposed to explore the temporal features of the data with two autoregressive models. These models perform a tough cross-view prediction task by predicting the future of one view using the past of the other. We further maximize the agreement between the contexts of the autoregressive models by a contextual contrasting module. These components are illustrated in Fig.~\ref{Fig:overall}.
For the semi-supervised settings, our proposed CA-TCC starts with TS-TCC pretraining phase and includes three more phases to complete the semi-supervised training, as shown in Fig.~\ref{Fig:supervised_contrastive}.
In the next subsections, we will introduce each component in more details.

\subsection{Time-Series Data Augmentation}

Data augmentation is a key part in the success of contrastive learning methods \cite{simclr_paper,grill2020bootstrap}. Contrastive methods try to maximize the similarity among different views of the same sample while minimizing its similarity with other samples. It is thus important to design proper data augmentations for contrastive learning. Usually, contrastive learning methods use two (random) variants of the same augmentation~\cite{simclr_paper,pmlr-v136-mohsenvand20a}. Specifically, given a sample $x$, they produce two views $x_1$ and $x_2$ sampled from the same augmentation family $\mathcal{T}$, i.e., $x_1 \!\sim\! \mathcal{T}$ and $x_2 \!\sim\! \mathcal{T}$. 
However, we argue that producing views from different augmentations can improve the robustness of the learned representations. Consequently, we propose two separate augmentations, such that one augmentation is weak and the other is strong. 

Our framework uses strong augmentation to enable the \textit{tough} cross-view prediction task in the next module, which helps in learning robust representations about the data. The weak augmentation aims to add some small variations to the signal without affecting its characteristics or making major changes in its shape. 
We include both types of augmentation to introduce variations of data, which increases the model's generalization ability to perform well in the unseen test set. We discuss more details about the augmentation selection in Section \ref{sec:aug_selec}.

Formally, for each input sample $x$, we denote its strongly augmented view as $x^s$, and its weakly augmented view as $x^w$, where $x^s \!\sim\! \mathcal{T}_s$ and $x^w \!\sim\! \mathcal{T}_w$.
These views are then passed to the encoder to extract their high dimensional latent representations. In particular, the encoder has a 3-block convolutional architecture as proposed in \cite{wang2017time}. For an input $\mathbf{x}$, the encoder maps $\mathbf{x}$ into a high-dimensional latent representation $\mathbf{z} = f_{enc}(\mathbf{x})$.
We define $\mathbf{z} = [z_1, z_2, \dots z_{T}]$, where $T$ is the total timesteps, $z_i \in \mathbb{R}^{d}$, where $d$ is the feature length.
Thus, we get $\mathbf{z}^s$ for the strong augmented views, and $\mathbf{z}^w$ for the weak augmented views, which are then fed into the temporal contrasting module.

\subsection{Temporal Contrasting}
The Temporal Contrasting module deploys a contrastive loss to extract temporal features in the latent space with an autoregressive model.
Given the latent representations $\mathbf{z}$, the autoregressive model $f_{ar}$ summarizes all $\mathbf{z}_{\leq t}$ into a context vector $c_t = f_{ar}(\mathbf{z}_{\leq t}),~ c_t \in \mathbb{R}^{h}$, where $h$ is the hidden dimension of $f_{ar}$.
The context vector $c_t$ is then used to predict the timesteps from $z_{t+1}$ until $z_{t+k}$ $(1<k\leq K)$. To predict future timesteps, we use a log-bilinear model that would preserve the mutual information between the input $x_{t+k}$ and $c_t$, such that $f_k(x_{t+k}, c_t) = exp((\mathcal{W}_k (c_t))^T z_{t+k})$, where $\mathcal{W}_k$ is a linear function that maps $c_t$ back into the same dimension as $z$, i.e., $\mathcal{W}_k : \mathbb{R}^{h \rightarrow d}$.

\begin{figure}
\centering
\includegraphics[width=\columnwidth]{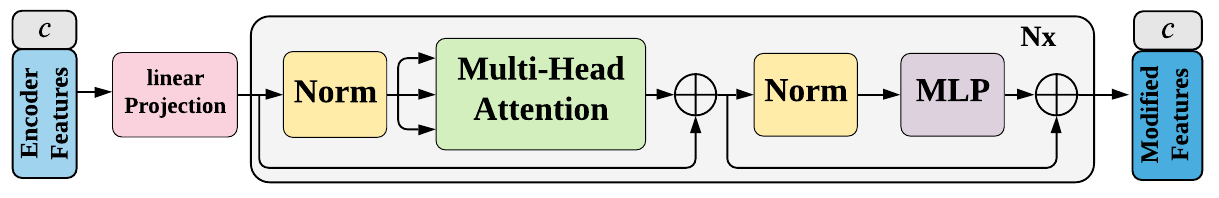}
\caption{Architecture of the Transformer model used in the Temporal Contrasting module. We first attach the classification token $c$ to the input and pass it through the different Transformer model layers. Eventually,  we split the token from the output modified features and use it in the next Contextual Contrasting module.}
\label{Fig:transformer}
\end{figure}

In our approach, the strong augmentation generates $c_t^s$ and the weak augmentation generates $c_t^w$. We propose a tough cross-view prediction task by using the context of the strong augmentation $c_t^s$ to predict the future timesteps of the weak augmentation $z_{t+k}^w$ and vice versa. The contrastive loss tries to maximize the dot product between the predicted representation and the true one of the same sample while minimizing the dot product with the other samples $\mathcal{N}_{t, k}$ within the mini-batch.
Accordingly, we calculate the two losses $ \mathcal{L}_{TC}^s$ and $ \mathcal{L}_{TC}^w$ as follows:
\begin{equation}
    \mathcal{L}_{TC}^s=-\frac{1}{K} \sum_{k=1}^{K} \log \frac{\exp ((\mathcal{W}_k (c_t^s))^T z_{t+k}^{w})}{\sum_{n \in \mathcal{N}_{t, k}} \exp ((\mathcal{W}_k (c_t^s))^T z_{n}^w)}
\end{equation}
\begin{equation}
    \mathcal{L}_{TC}^w=-\frac{1}{K} \sum_{k=1}^{K} \log \frac{\exp ((\mathcal{W}_k (c_t^w))^T z_{t+k}^{s})}{\sum_{n \in \mathcal{N}_{t, k}} \exp ((\mathcal{W}_k (c_t^w))^T z_{n}^s)}
\end{equation}

\begin{figure*}
\centering
\includegraphics[width=0.7\textwidth]{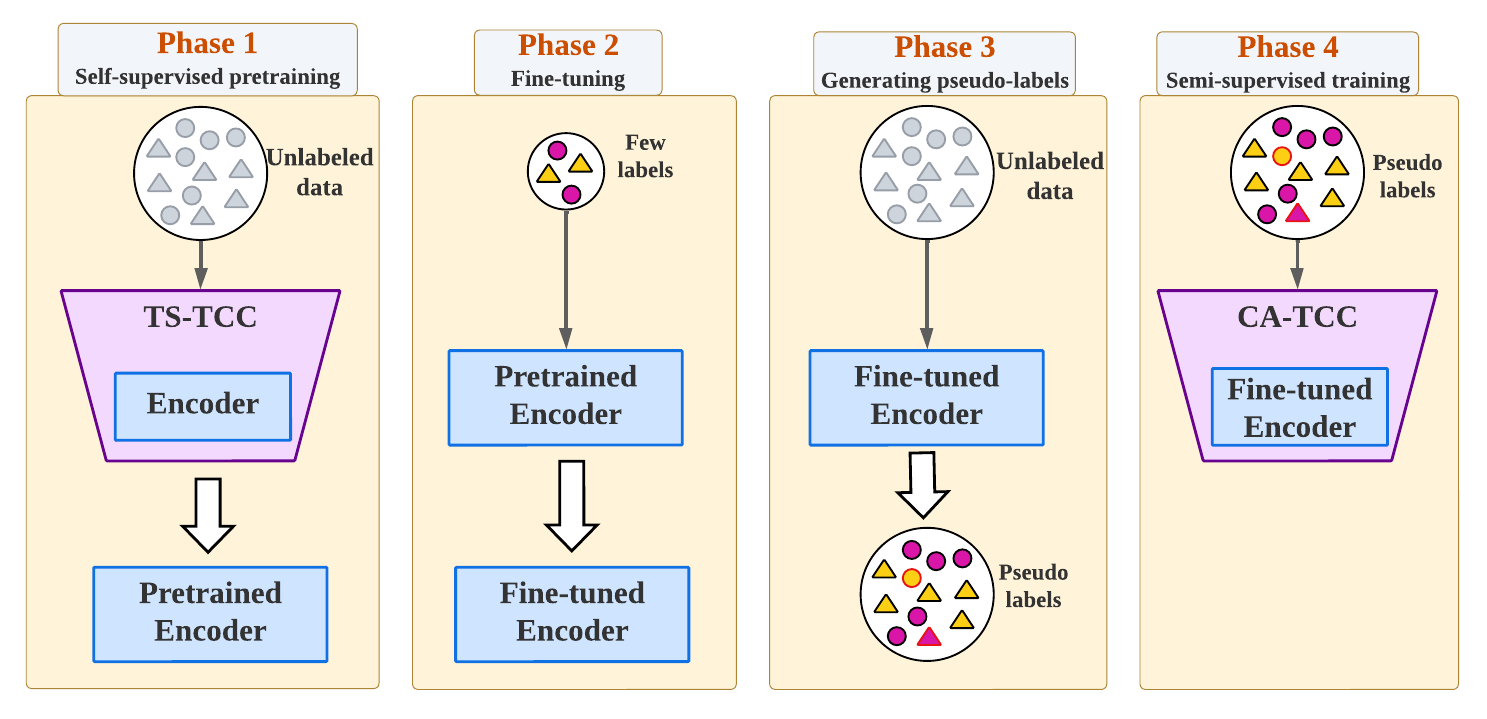}
\caption{The four phases for CA-TCC semi-supervised training. In Phase~1, TS-TCC is trained with fully unlabeled data. Next, we use the available few labeled samples to fine-tune the pretrained encoder in Phase~2. Following that, in Phase~3, we generate pseudo labels for the unlabeled data with the fine-tuned TS-TCC encoder. Finally, Phase~4 includes deploying the pseudo-labeled data in the semi-supervised training.}
\label{Fig:supervised_contrastive}
\end{figure*}

We use Transformer as the autoregressive model because of its efficiency and speed \cite{NIPS2017_3f5ee243}. The architecture of the Transformer model is shown in Fig.~\ref{Fig:transformer}. It mainly consists of multi-headed attention (MHA) followed by a Multilayer Perceptron (MLP) block. The MLP block is composed of two fully-connected layers with a non-linearity ReLU function and dropout in between. Pre-norm residual connections, which can produce more stable gradients \cite{wang-etal-2019-learning}, are adopted in our Transformer. We stack $L$ identical layers to generate the final features. Inspired by the BERT model \cite{devlin2018bert}, we add a token $c \in \mathbb{R}^{h}$ to the input, whose state acts as a representative context vector in the output.
The operation of the Transformer starts by applying the features $\mathbf{z}_{\leq t}$ to a linear projection $\mathcal{W}_{Tran}$ layer that maps the features into the hidden dimension, i.e., $\mathcal{W}_{Tran} : \mathbb{R}^{d \rightarrow h}$. The output of this linear projection is then sent to the Transformer i.e., $\tilde{\mathbf{z}} = \mathcal{W}_{Tran} (\mathbf{z}_{\leq t}),~~\tilde{\mathbf{z}} \in \mathbb{R}^{h}$.
Next, we attach the context vector into the feature vector $\tilde{\mathbf{z}}$ such that the input features become $\psi_0 = [c; \tilde{\mathbf{z}}]$, where the subscript 0 denotes being the input to the first layer.
Next, we pass $\psi_0$ through Transformer layers as in the following equations:
\begin{align}
    \tilde{\psi_l} &= \operatorname{MHA}(\operatorname{Norm}(\psi_{l-1})) + \psi_{l-1}, && 1 \leq l \leq L; \label{eq:msa_apply} \\
    \psi_l &= \operatorname{MLP}(\operatorname{Norm}(\tilde{\psi_l})) + \tilde{\psi_l}, && 1 \leq l \leq L.  \label{eq:mlp_apply}
\end{align}
Finally, we re-attach the context vector from the final output such that $c_t = \psi_L^0$.
This context vector will be the input of the following contextual contrasting module.

\subsection{Contextual Contrasting}
\label{sec:cc}
We further propose a contextual contrasting module that aims to learn more discriminative representations. It starts with applying a non-linear transformation to the contexts using a non-linear projection head as in \cite{simclr_paper}. The projection head maps the contexts into the space where the contextual contrasting is applied. 

Given a batch of $N$ input samples, we will have two contexts for each sample from its two augmented views and thus have $2N$ contexts. 
For a context $c_t^i$, we denote $c_t^{i^+}$ as the positive sample of $c_t^{i}$ that comes from the other augmented view of the same input, and hence, ($c_t^i, c_t^{i^+}$) is considered to be a positive pair.  
Meanwhile, the remaining ($2N-2$) contexts from other inputs within the same batch are considered as the negative samples of $c_t^i$, i.e., $c_t^i$ can form ($2N-2$) negative pairs with its negative samples. Therefore, we can derive a contextual contrasting loss to maximize the similarity between the sample and its positive pair, while minimizing its similarity with the negative samples within the mini-batch. As such, the final representations can be discriminative. 

Eq.~\ref{eqn:lcc} defines the contextual contrasting loss function $\mathcal{L}_{CC}$. Given a context $c_t^i$, we divide its similarity with its positive sample $c_t^{i^+}$ by its similarity with all the other ($2N - 1$) samples, including the positive pair and ($2N - 2$) negative pairs, to normalize the loss. 

\begin{align}
    &\ell(i,i^+) = - \log \frac{\exp \left(\operatorname{sim}\left(\boldsymbol{c}_t^{i}, \boldsymbol{c}_t^{i^+}\right) / \tau\right)}{\sum_{m=1}^{2 N} \mathbbm{1}_{[m \neq i]} \exp \left(\operatorname{sim}\left(\boldsymbol{c}_t^{i}, \boldsymbol{c}_t^{m}\right) / \tau\right)},\\
    &\mathcal{L}_{CC} = \frac{1}{2N}\sum_{k=1}^{2N}~\left[\ell(2k-1,2k) + \ell(2k,2k-1)\right],
    \label{eqn:lcc}
\end{align}
where  $sim(\bm u,\bm v) = \bm u^T \bm v/ \| \bm u\|\| \bm v\| $ denotes the dot product between $\ell_2$ normalized $\bm u$ and $\bm v$ (i.e., cosine similarity), $\mathbbm{1}_{[m \neq i]} \in \{0,1\}$ is an indicator function, evaluating to $1$ iff $m \neq i$, and $\tau$ is a temperature parameter.

The overall self-supervised loss is the combination of the two temporal contrasting losses and the contextual contrasting loss as follows.
\begin{equation}
    \mathcal{L}_{unsup} = \lambda_1 \cdot (\mathcal{L}_{TC}^s + \mathcal{L}_{TC}^w ) + \lambda_2 \cdot \mathcal{L}_{CC},
    \label{eq:overall}
\end{equation}
where $\lambda_1$ and $\lambda_2$ are fixed scalar hyperparameters denoting the relative weight of each loss.

\begin{figure}[!tb]
\centering
\includegraphics[width=\columnwidth]{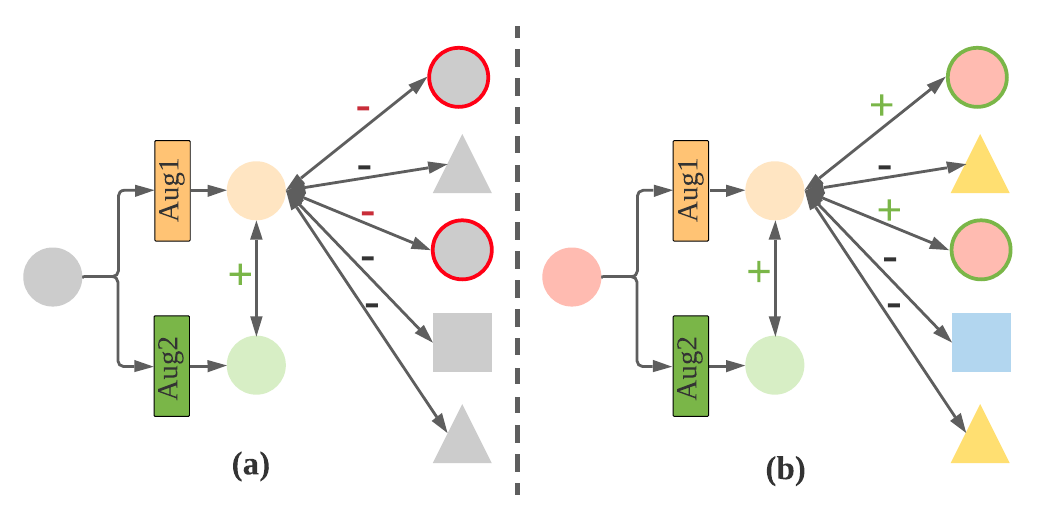}
\caption{(a) Unsupervised contrasting vs. (b) supervised contrasting. The unsupervised contrasting forms positive pairs only from the augmented views of the sample, which may deteriorate the performance. In contrast, supervised contrasting considers the semantics of the data and forms positive pairs among all the samples having the same class label. In CA-TCC, we deploy the pseudo-labeled data to apply the supervised contrasting in the semi-supervised training.}
\label{Fig:unsup_vs_sup}
\end{figure}

\subsection{Class-Aware TS-TCC}
We propose a second variant of our framework, called the Class-Aware TS-TCC (CA-TCC), to operate in semi-supervised settings. In this variant, we attempt to exploit the available few labeled data to further improve the representation learned by TS-TCC.
Our semi-supervised learning framework is performed in four phases, as shown in Fig.~\ref{Fig:supervised_contrastive}. In Phase~1, we start with a randomly-initialized encoder and used it in TS-TCC self-supervised pretraining. In Phase~2, we fine-tune the pretrained encoder with the few labeled data. In Phase~3, we deploy the fine-tuned encoder to create pseudo labels for the entire unlabeled dataset. Finally, in Phase~4, we train the class-aware semi-supervised framework, i.e., CA-TCC, with the pseudo labels.

In the class-aware semi-supervised training, we replace the unsupervised contextual contrasting with a supervised contextual contrasting module. This module benefits from the pseudo-labeled data to train a supervised contrastive loss~\cite{khosla2020supervised}.
This loss considers samples with the same class label as positive pairs, while samples from different classes are considered as negative pairs. This is different from the unsupervised contrastive loss that only forms positive pairs from the augmented views of the sample, as depicted in Fig.~\ref{Fig:unsup_vs_sup}. This difference can affect the performance of the model, as the unsupervised contrastive loss may treat samples having a similar class as negative pairs.

Formally, assuming that the dataset consists of $N$ labeled samples $\{\mathbf{x}_k, y_k\}_{k=1 \dots N}$, then after applying augmentations, the dataset becomes of 2$N$ samples, $\{\mathbf{\hat{x}}_l, \hat{y}_l\}_{l=1 \dots 2N}$ such that $\mathbf{\hat{x}}_{2k}$ and $\mathbf{\hat{x}}_{2k-1}$ are the two views of $\mathbf{x}_k$, and similarly $y_{k} = \hat{y}_{2k} = \hat{y}_{2k-1}$. 
Also assuming that $i \in I \equiv \{1 \dots 2N\}$ represents the index of an arbitrary augmented sample, and $A(i) \equiv I \setminus \{i\}$, the supervised contextual contrasting loss can be expressed as:
\begin{align}
    &\mathcal{L}_{SCC}= \sum_{i \in I} \frac{1}{|P(i)|} \sum_{p \in P(i)} \ell(i,p), \\
    &P(i) = \{ p \in A(i) :  \hat{y}_p = \hat{y}_i\},
    \label{eqn:lscc}
\end{align}
where $P(i)$ is the set of indices of all samples with the same class as the sample $\mathbf{\hat{x}}_{i}$ in the batch, and ($i, p$) for any $p \in P(i)$ is thus a positive pair. $|P(i)|$ is the cardinality of $P(i)$.
Thus, the overall loss can be expressed as:
\begin{equation}
    \mathcal{L}_{semi} = \lambda_3 \cdot (\mathcal{L}_{TC}^s + \mathcal{L}_{TC}^w ) + \lambda_4 \cdot \mathcal{L}_{SCC},
    \label{eq:overall_2}
\end{equation}
where $\lambda_3$ and $\lambda_4$ are fixed scalar hyperparameters denoting the relative weight of each loss.

\section{Experimental Setup}
\subsection{Datasets}
To comprehensively evaluate our proposed models, we adopted ten publicly available real-world datasets that cover different time-series applications. Additionally, we investigated the transferability of our learned features on a fault diagnosis dataset.

\subsubsection{Human Activity Recognition}
We employed UCI HAR dataset\footnote{\url{https://archive.ics.uci.edu/ml/datasets/human+activity+recognition+using+smartphones}} \cite{anguita2013public}, which contains sensor readings for 30 subjects performing 6 activities (i.e., walking, walking upstairs, downstairs, standing, sitting, and lying down). The data were collected by a mounted Samsung Galaxy S2 device on the users' waist, with a sampling rate of 50 Hz.

\subsubsection{Sleep Stage Classification}
We adopted Sleep-EDF dataset\footnote{\url{https://physionet.org/physiobank/database/sleep-edfx/}} \cite{goldberger2000physiobank}, which includes whole-night polysomnography (PSG) sleep recordings for 20 subjects. In particular, each recording contains two electroencephalogram (EEG) channels namely Fpz-Cz and Pz-Oz, with a sampling rate of 100 Hz.
In this work, we used a single EEG channel (i.e., Fpz-Cz), following previous studies \cite{supratak2017deepsleepnet,emadeldeen_attnSleep}.
Sleep stage classification refers to classifying the input EEG signal into one of five classes:  Wake (W), Non-rapid eye movement (N1, N2, N3), and Rapid Eye Movement (REM).

\subsubsection{Epilepsy Seizure Prediction}
The Epileptic Seizure Recognition dataset\footnote{\href{https://archive.ics.uci.edu/ml/datasets/Epileptic+Seizure+Recognition}{archive.ics.uci.edu/ml/datasets/Epileptic+Seizure+Recognition}} \cite{PhysRevE.64.061907} consists of EEG recordings from 500 subjects, where the brain activity was recorded for each subject for 23.6 seconds. The original dataset is labeled with five classes, but as four of them do not include epileptic seizures, we merged them into one class and treat it as a binary classification problem.

\subsubsection{Fault Diagnosis (FD)}
Fault Diagnosis dataset\footnote{\url{https://mb.uni-paderborn.de/en/kat/main-research/datacenter/bearing-datacenter/data-sets-and-download}} \cite{lessmeier2016condition} was collected from sensor readings of bearing machine under four different working conditions.
Each working condition can be considered as a separate domain as it has different characteristics from the other working conditions (e.g., rotational speed and load torque) \cite{ragab2020adversarial}.
Each domain has three classes; two fault classes (i.e., inner fault and outer fault) and one healthy class. 
We use this dataset to conduct the transferability experiment only and show the effectiveness of our method in transfer learning scenarios.


\subsubsection{UCR repository datasets}
We included seven UCR datasets\footnote{\label{ts_url}\url{https://timeseriesclassification.com/dataset.php}} in our experiments. The selected datasets are suitable for our few labels experiments, and they satisfy the condition that 1\% of samples allow all the classes to be present during the training phase.
The included datasets are: Wafer, FordA, FordB, PhalangesOutlinesCorrect (POC), ProximalPhalanxOutlineCorrect (PPOC), StarLightCurves, and ElectricDevices. We included a detailed discussion about each of these datasets in the supplementary materials.

\begin{table}[!bth]

\centering
\caption{A brief description of the 10 datasets used in our experiments. For the FD dataset, we mentioned the settings of one working condition, as they are the same for the four working conditions.}
\resizebox{0.48\textwidth}{!}{
\begin{tabular}{@{}l|ccccc@{}}
\toprule
Dataset & \# Train & \# Test & Length & \# Channel & \# Class \\ \midrule
HAR & 7,352 & 2,947 & 128 & 9 & 6 \\
Sleep-EDF & 25,612 & 8,910 & 3,000 & 1 & 5 \\
Epilepsy & 9,200 & 2,300 & 178 & 1 & 2 \\ 
FD & 8,184 & 2,728 & 5,120 & 1 & 3 \\ 
Wafer & 1,000 & 6,174 & 152 & 1 & 2 \\
FordA & 1,320 & 3,601 & 500 & 1 & 2 \\
FordB & 810 & 3,636 & 500 & 1 & 2 \\
POC & 1,800 & 858 & 80 & 1 & 2 \\
PPOC & 600 & 291 & 80 & 1 & 2 \\
StarLightCurves & 1,000 & 8,236 & 1,024 & 1 & 3 \\
ElectricDevices & 8,926 & 7,711 & 96 & 1 & 7 \\

\bottomrule
\end{tabular}
}

\label{tab:data}
\end{table}

Table \ref{tab:data} summarizes the statistical details of each dataset, i.e., the number of training samples (\# Train) and testing samples (\# Test), the length of the sample, the number of sensor channels (\# Channel) and the number of classes (\# Class).

\begin{table*}[!htb]
    \centering
    \caption{Results of fine-tuning SELF-SUPERVISED pretrained models with 1\% and 5\% of labels. \textbf{Best results} across each row are in bold, while the \underline{second-best results} are underlined.}
    \resizebox{\textwidth}{!}{
    \begin{tabular}{@{}l|cc|cc|cc|cc|cc|cc@{}}
    \toprule
    \multicolumn{13}{c}{\textbf{1\% of labeled data}}  \\ \midrule
     & \multicolumn{2}{c|}{Random Init} & \multicolumn{2}{c|}{Supervised} & \multicolumn{2}{c|}{SSL-ECG} & \multicolumn{2}{c|}{CPC} & \multicolumn{2}{c|}{SimCLR} & \multicolumn{2}{c}{TS-TCC} \\ 
     \midrule
     
     Datasets & Accuracy & MF1-score & Accuracy & MF1-score & Accuracy & MF1-score & Accuracy & MF1-score & Accuracy & MF1-score & Accuracy & MF1-score \\ \midrule
            
            HAR & 39.8$\pm$3.8 & 34.7$\pm$5.0 & 44.9$\pm$6.7 & 41.0$\pm$6.7 & 60.0$\pm$4.0 & 54.0$\pm$6.0 & 65.4$\pm$1.4 & 63.8$\pm$1.7 & \underline{65.8$\pm$0.7} & \underline{64.3$\pm$0.9} & \textbf{70.5$\pm$0.3} & \textbf{69.5$\pm$0.5} \\
            
            Sleep-EDF & 20.5$\pm$1.0 & 20.7$\pm$1.4 & 34.1$\pm$0.3 & 30.8$\pm$0.8 & 70.9$\pm$0.8 & 61.6$\pm$0.7 & \underline{74.7$\pm$0.2} & \underline{68.7$\pm$0.0} & 58.0$\pm$1.0 & 56.9$\pm$0.6 & \textbf{75.8$\pm$0.4} & \textbf{70.0$\pm$0.2} \\
            
            Epilepsy & 70.3$\pm$2.1 & 66.2$\pm$2.6 & 76.1$\pm$0.7 & 74.8$\pm$0.4 & \underline{89.3$\pm$0.4} & \underline{86.0$\pm$0.3} & 88.9$\pm$1.1 & 85.8$\pm$0.3 & 88.3$\pm$1.5 & 84.0$\pm$1.0 & \textbf{91.2$\pm$0.5} & \textbf{89.2$\pm$0.2} \\
            
            
            Wafer & 90.6$\pm$1.6 & 58.1$\pm$2.1 & 91.9$\pm$1.3 & 67.6$\pm$9.2 & 93.4$\pm$0.5 & 76.1$\pm$2.4 & \underline{93.5$\pm$0.4} & \underline{78.4$\pm$1.5} & \textbf{93.8$\pm$0.2} & \textbf{78.5$\pm$1.1} & 93.2$\pm$0.8 & 76.7$\pm$4.6  \\ 
            
            FordA & 50.6$\pm$2.1 & 36.6$\pm$4.2 & 56.4$\pm$1.6 & 54.4$\pm$3.5 & 67.9$\pm$8.8 & 66.2$\pm$9.5 & \underline{75.8$\pm$1.4} & \underline{75.2$\pm$1.8} & 55.9$\pm$3.7 & 55.7$\pm$3.8 & \textbf{80.6$\pm$2.0} & \textbf{80.0$\pm$2.4} \\ 
            
            FordB & 52.5$\pm$1.8 & 47.5$\pm$6.4 & 51.9$\pm$2.6 & 48.0$\pm$3.6 & 64.4$\pm$6.2 & 60.5$\pm$7.9 & \underline{66.8$\pm$3.1} & \underline{65.0$\pm$3.9} & 50.9$\pm$1.3 & 49.8$\pm$2.2 & \textbf{72.7$\pm$0.9} & \textbf{71.9$\pm$1.0} \\ 
            
            POC & 61.4$\pm$0.0 & 38.3$\pm$0.0 & 62.0$\pm$0.8 & 40.0$\pm$2.1 & 62.5$\pm$1.8 & 41.2$\pm$4.9 & \textbf{64.8$\pm$1.0} & \textbf{48.2$\pm$2.9} & 61.5$\pm$0.1 & 38.4$\pm$0.3 & \underline{63.8$\pm$0.5} & \underline{48.1$\pm$0.9}  \\ 
            
            PPOC & 68.4$\pm$0.0 & 40.6$\pm$0.0 & \textbf{64.3$\pm$0.7} & \textbf{64.2$\pm$0.7} & 49.8$\pm$7.3 & 37.6$\pm$8.4 & 63.3$\pm$0.7 & 63.0$\pm$0.8 & 37.6$\pm$5.1 & 32.8$\pm$7.7 & \underline{63.4$\pm$0.3} & \underline{63.1$\pm$0.4} \\ 
            
            StarLightCurves & 83.9$\pm$1.6 & 65.4$\pm$3.6 & 78.8$\pm$0.9 & 71.4$\pm$0.1 & 78.3$\pm$0.9 & 72.0$\pm$0.8 & \underline{80.8$\pm$1.4} & \underline{74.4$\pm$0.6} & 80.6$\pm$0.6 & 71.6$\pm$0.2 & \textbf{86.0$\pm$0.4} & \textbf{79.2$\pm$0.7} \\ 
            
            ElectricDevices & 50.8$\pm$4.4 & 41.9$\pm$3.6 & 57.8$\pm$1.1 & 47.5$\pm$1.0 & 60.1$\pm$4.1 & 50.0$\pm$4.9 & 59.3$\pm$4.1 & 48.9$\pm$6.7 & \underline{62.5$\pm$1.1} & \underline{51.2$\pm$0.5} & \textbf{63.6$\pm$1.2} & \textbf{56.4$\pm$0.6}  \\

            \midrule
    
            Average & 58.8 & 45.0 & 61.8 & 54.0 & 69.7 & 60.5 & \underline{73.3} & \underline{67.1} & 65.5 & 58.3 & \textbf{76.1} & \textbf{70.4} \\
            
        \toprule
         \multicolumn{13}{c}{\textbf{5\% of labeled data}}  \\ 
        \toprule
        
        HAR & 49.6$\pm$2.5 & 45.8$\pm$2.0 & 52.8$\pm$1.5 & 50.9$\pm$0.2 & 63.7$\pm$5.3 & 58.6$\pm$7.4 & 75.4$\pm$2.1 & 74.7$\pm$2.5 & \underline{75.8$\pm$1.4} & \underline{74.9$\pm$1.5} & \textbf{77.6$\pm$1.8} & \textbf{76.7$\pm$1.7} \\
        
        Sleep-EDF & 22.8$\pm$2.8 & 22.8$\pm$2.2 & 60.5$\pm$3.9 & 54.8$\pm$5.5 & 73.4$\pm$0.5 & 63.7$\pm$0.1 & \underline{76.3$\pm$0.4} & \underline{70.5$\pm$0.3} & 64.2$\pm$1.0 & 61.9$\pm$0.8 & \textbf{77.0$\pm$0.6} & \textbf{70.9$\pm$0.5} \\
        
        Epilepsy & 75.5$\pm$3.6 & 70.5$\pm$3.3 & 83.4$\pm$0.7 &  80.4$\pm$0.7 & 92.8$\pm$0.2 & 89.0$\pm$0.3 & \underline{92.8$\pm$0.3} & \underline{90.2$\pm$0.5} & 91.3$\pm$0.5 & 89.2$\pm$1.0 & \textbf{93.1$\pm$0.3} & \textbf{93.7$\pm$0.6} \\
        
        
        Wafer & 91.2$\pm$1.2 & 65.5$\pm$8.2 & 94.6$\pm$0.3 & 83.9$\pm$0.6 & \textbf{94.9$\pm$0.3} & \textbf{84.5$\pm$0.7} & 92.5$\pm$0.4 & 79.4$\pm$0.8 & \underline{94.8$\pm$0.2} & \underline{83.3$\pm$0.6} & 93.2$\pm$0.4 & 81.2$\pm$0.7 \\

        FordA & 54.4$\pm$2.4 & 50.5$\pm$7.6 & 54.5$\pm$4.3 & 44.1$\pm$8.0 & 73.6$\pm$1.2 & 70.7$\pm$1.5 & \underline{86.5$\pm$1.9} & \underline{86.5$\pm$1.9} & 69.6$\pm$1.3 & 68.9$\pm$1.7 & \textbf{89.9$\pm$0.1} & \textbf{89.9$\pm$0.1}  \\ 
        
        FordB & 51.3$\pm$3.2 & 48.2$\pm$5.3 & 60.5$\pm$2.8 & 58.8$\pm$3.7 & 71.7$\pm$3.1 & 69.8$\pm$3.8 & \textbf{86.3$\pm$0.8} & \textbf{86.2$\pm$0.8} & 63.0$\pm$3.0 & 60.7$\pm$4.2 & \underline{86.1$\pm$1.5} & \underline{85.9$\pm$1.6}  \\

        POC & 61.6$\pm$0.3 & 38.8$\pm$1.0 & 61.4$\pm$0.0 & 38.3$\pm$0.0 & \underline{62.9$\pm$0.3} & \underline{43.3$\pm$1.4} & \textbf{66.9$\pm$2.6} & \textbf{44.3$\pm$8.4} & 62.7$\pm$1.1 & 42.4$\pm$4.0 & 62.6$\pm$1.1 & 42.6$\pm$3.0  \\

        PPOC & 64.1$\pm$2.8 & 57.4$\pm$9.8 & 69.1$\pm$2.4 & 62.2$\pm$6.4 & 68.8$\pm$0.0 & 40.7$\pm$0.0 & \underline{71.5$\pm$3.4} & \underline{63.9$\pm$1.5} & 48.0$\pm$2.3 & 42.9$\pm$2.1 & \textbf{72.1$\pm$4.4} & \textbf{64.2$\pm$3.7} \\ 
        
        StarLightCurves & 74.2$\pm$1.4 & 69.8$\pm$4.1 & 81.8$\pm$0.8 & 71.4$\pm$4.1 & 82.6$\pm$1.3 & 74.5$\pm$1.3 & \textbf{89.1$\pm$1.0} & \textbf{84.5$\pm$0.8} & 84.2$\pm$1.3 & 74.0$\pm$2.3 & \underline{89.6$\pm$0.2} & \underline{82.7$\pm$0.9} \\ 
        
        ElectricDevices & 57.4$\pm$1.2 & 52.3$\pm$1.0 & 59.7$\pm$0.7 & 55.6$\pm$0.8 & 63.7$\pm$0.8 & 56.1$\pm$2.2 & 62.4$\pm$0.6 & 58.1$\pm$0.6 & \underline{63.9$\pm$1.2} & \underline{58.6$\pm$0.6} & \textbf{65.1$\pm$0.3} & \textbf{59.2$\pm$0.4} \\ 
        
        \midrule

        Average & 60.2 & 52.2 & 67.8 & 60.0 & 74.8 & 65.1 & \underline{80.0} & \underline{73.8} & 71.8 & 65.7 & \textbf{80.6} & \textbf{74.7}  \\
    
    \bottomrule
    \end{tabular}
    }
    \label{tbl:ft_self_ucr}
\end{table*}

\subsection{Implementation Details}
We split the data into 60\%, 20\%, and 20\% for training, validation, and testing with considering subject-wise split for the Sleep-EDF dataset to avoid overfitting.
Experiments were repeated 5 times with 5 different seeds, and we reported the performance with mean and standard deviation.
The pretraining and downstream tasks were done for 40 epochs, as we noticed that the performance did not improve with further training.
We applied a batch size of 128 (which was reduced in \textit{few-labeled data} experiments as data size may be less than 128).
We used Adam optimizer with a learning rate of 3e-4, weight decay of 3e-4, $\beta_1 = 0.9$, and $\beta_2 = 0.99$.
For the strong augmentation, we set the number of segments $M$ as $M_{Ep}$=12, $M_{EDF}$=20, and $M$=10 for all other datasets, (see Section~\ref{sec:aug_selec}), while for the weak augmentation, we set the scaling ratio to 2.
In TS-TCC, we set $\lambda_1=1$, while we achieved good performance when $\lambda_2$ is around 1. Particularly, we set $\lambda_2$ as 0.7 in our experiments. For CA-TCC, was set $\lambda_3=0.01$ and $\lambda_4=0.7$ (see Section~\ref{sec:sens_analysis}).
In the Transformer, we set the $L=4$, and the number of heads as 4. We tuned $h \in \{32, 50, 64, 100, 128, 200, 256\}$ and set $h_{EDF}$=64 and $h$=100 for all other datasets. We also set its dropout to 0.1.
In contextual contrasting, we set $\tau=0.2$.
Lastly, we built our model using PyTorch 1.7 and trained it on NVIDIA GeForce RTX 2080 Ti GPU.

\section{Experimental Results}
We conduct several experiments to demonstrate the efficacy of our proposed TS-TCC and CA-TCC models.
We measure the performance using two metrics namely the accuracy and the macro-averaged F1-score (MF1) to better verify performance on the imbalanced datasets. 
These metrics are defined as follows.
\begin{align}
    &\text{Accuracy} = \frac{\sum_{i=1}^{\mathcal{K}}TP_i}{N}, \label{equ:acc}\\
    &\text{MF1-score} = \frac{1}{\mathcal{K}} \sum_{i=1}^{\mathcal{K}} \frac{2 \times Precision_i \times Recall_i}{Precision_i + Recall_i}, \label{equ:f1}
\end{align}
where $Precision_i = \frac{TP_i}{TP_i + FP_i}$, and $ Recall_i = \frac{TP_i}{TP_i + FN_i} $. $TP_i, ~FP_i$, and $FN_i$ denote the True Positives, False Positives, and False Negatives for the $i$-th class respectively, $N$ is the total number of samples, and $\mathcal{K}$ is the number of classes in the dataset.
Although the results of these two equations are expected to be in a range from 0 to 1, we report them in this paper as a percentage.

\begin{figure*}[!b]
     \centering
     \begin{subfigure}[b]{0.3\textwidth}
         \centering
         \includegraphics[width=\textwidth]{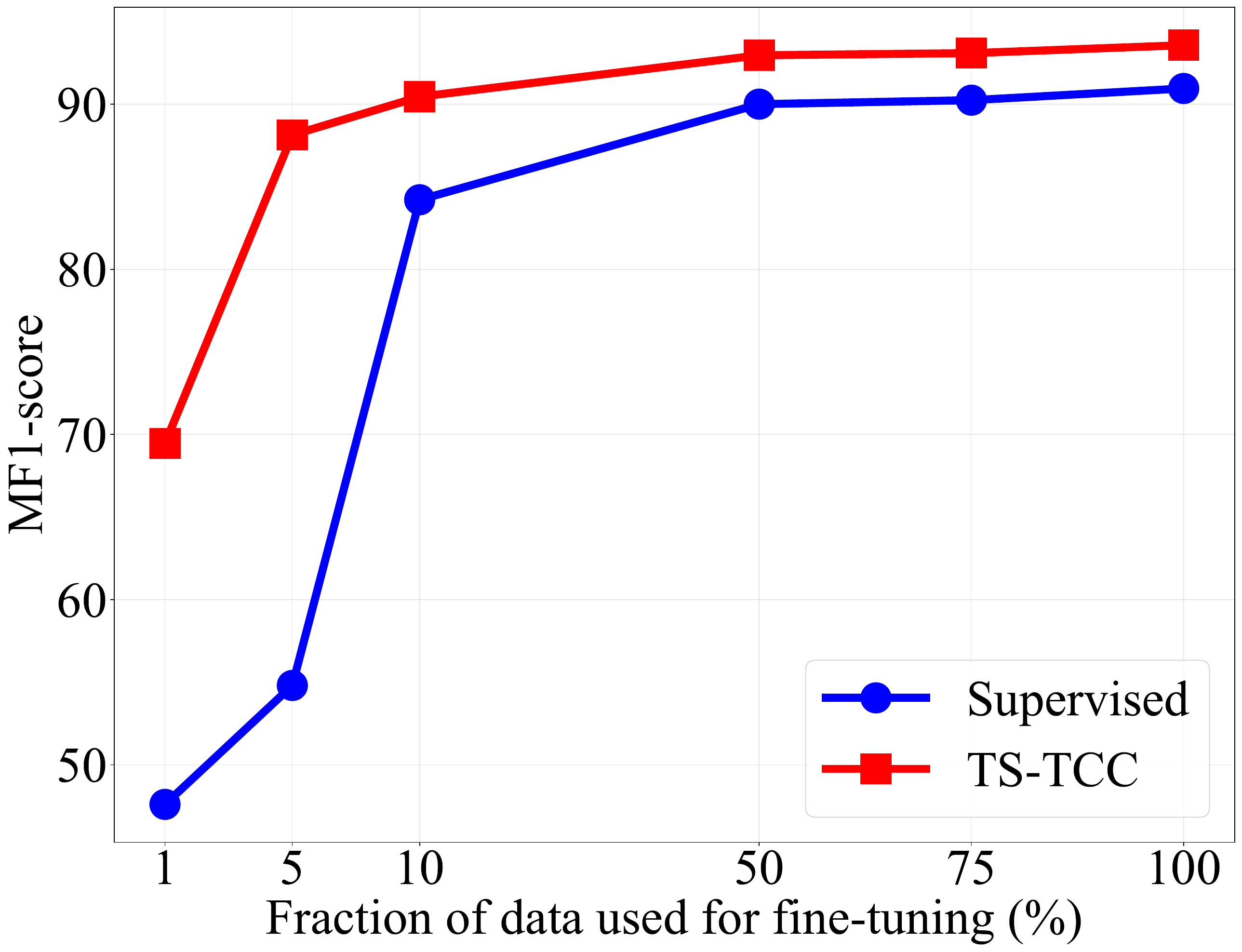}
         \caption{HAR dataset}
         \label{fig:semi1}
     \end{subfigure}
     \hfill
     \begin{subfigure}[b]{0.3\textwidth}
         \centering
         \includegraphics[width=\textwidth]{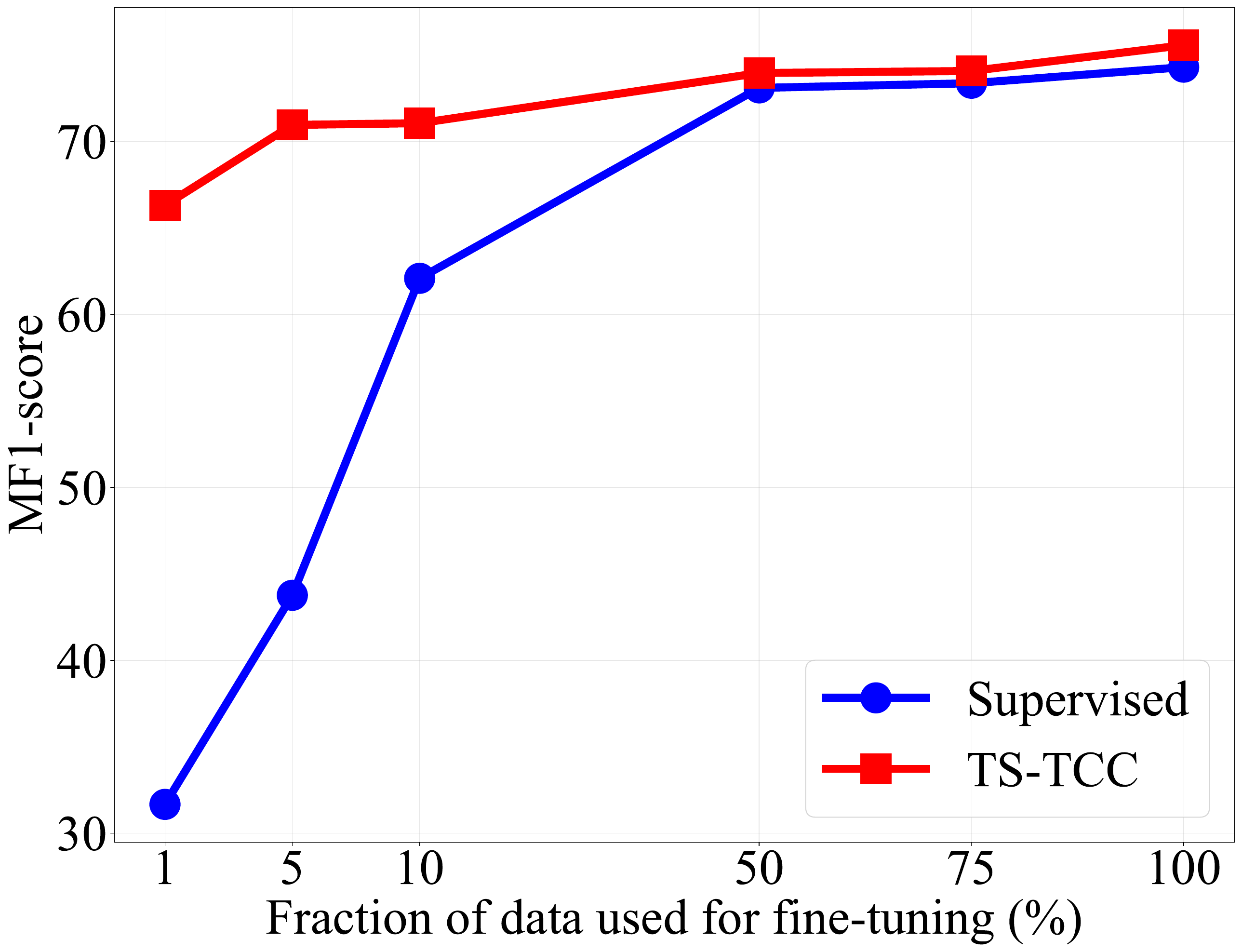}
         \caption{Sleep-EDF dataset}
         \label{fig:semi2}
     \end{subfigure}
     \hfill
     \begin{subfigure}[b]{0.3\textwidth}
         \centering
         \includegraphics[width=\textwidth]{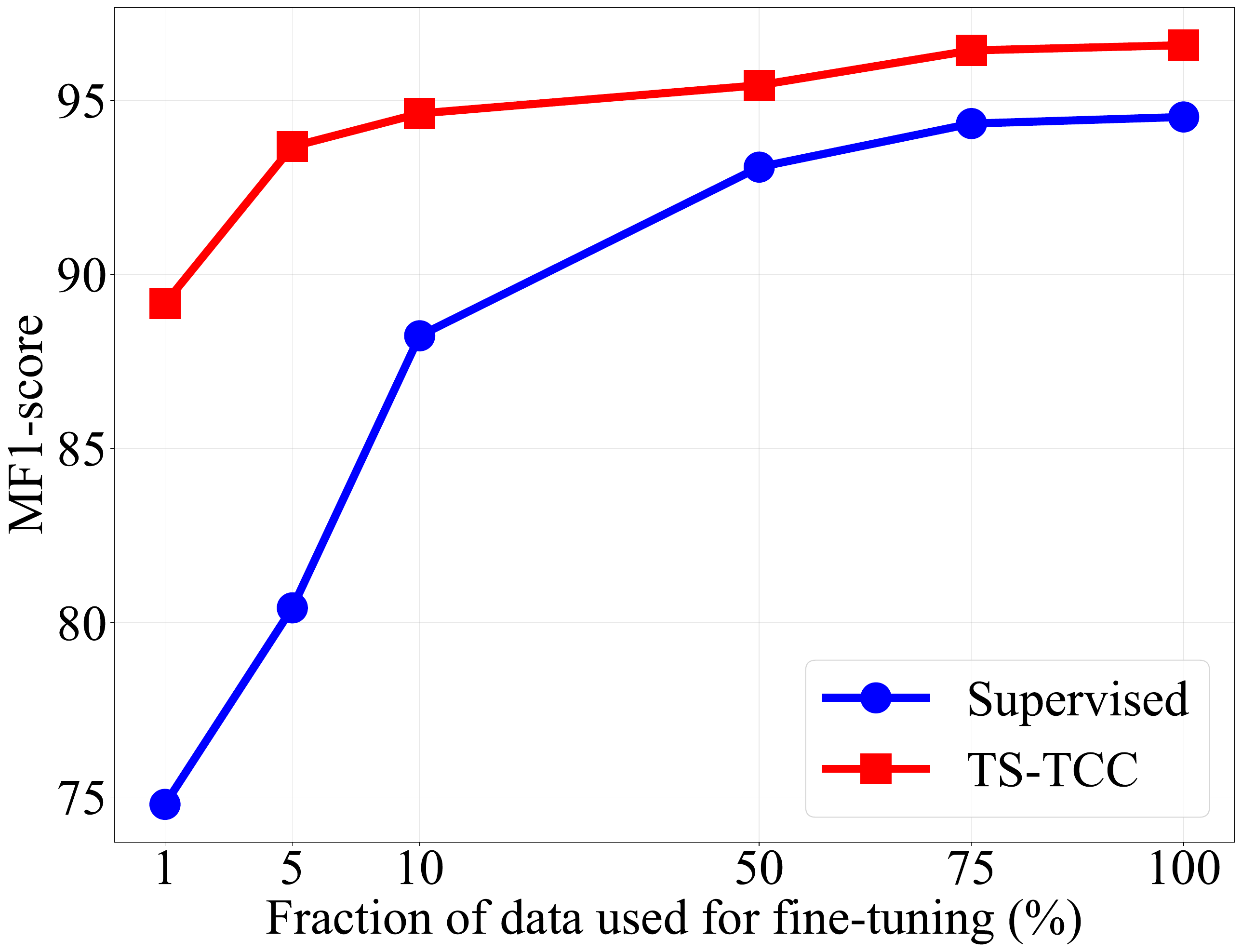}
         \caption{Epilepsy dataset}
         \label{fig:semi3}
     \end{subfigure}
        \caption{Comparison between supervised training vs. TS-TCC \textit{fine-tuning} on three datasets with different fractions of few labeled data in terms of MF1-score.}
        \label{Fig:few_labels}
\end{figure*}

\subsection{Results of Self-supervised TS-TCC Model}
\subsubsection{Comparison with Baseline Approaches}
We compare our proposed TS-TCC against the following baselines.

\begin{enumerate}
    \item \textbf{Random Initialization}: by training a linear classifier on top of a frozen and randomly initialized encoder, and this represents the lower bound.
    \item \textbf{Supervised}: supervised training of both: the encoder and the classifier.
    \item  \textbf{SSL-ECG}~\cite{ecg_emotion_rec}: an auxiliary-based self-supervised learning algorithm, based on classifying signal transformations.
    \item \textbf{CPC}~\cite{oord2018representation}: a contrastive self-supervised learning method that predicts future timesteps in the embedding space.
    \item \textbf{SimCLR}~\cite{simclr_paper}: a contrastive self-supervised learning method that deploys the unsupervised contrastive loss based on data augmentations.
\end{enumerate}
It is worth noting that, we use our time-series specific augmentations to pretrain SimCLR, as it was originally designed for images. 

To evaluate the performance of SSL-ECG, CPC, SimCLR, and TS-TCC, we first pretrain each of them with the fully unlabeled data, then use a portion of the labeled data to evaluate their performance.
We follow the standard \textit{linear evaluation} scheme~\cite{simclr_paper,oord2018representation}, in which we train a linear classifier (single fully connected layer) on top of a \textit{frozen} self-supervised pretrained encoder model.

Table~\ref{tbl:ft_self_ucr} shows the linear evaluation results of our approaches against the baseline methods using 1\% and 5\% of the labeled data. 
The results show that TS-TCC achieves the best overall performance across all 10 datasets in both the 1\% and 5\% label scenarios. Specifically, TS-TCC is the best performer on six datasets and the second best on two datasets with only a small margin behind the best methods. This demonstrates that our proposed cross-view prediction task can effectively improve representation learning for various time-series datasets. 
It is worth noting that contrastive methods (e.g., CPC, SimCLR, and our TS-TCC) generally achieve better results than the pretext-based method (i.e., SSL-ECG), which reflects the power of invariant features learned by contrastive methods.
Additionally, CPC shows better results than SimCLR, being the best on one dataset and the second best on five datasets. This highlights the importance of learning temporal dependencies in time-series data for better self-supervised representation learning, as implemented in CPC and TS-TCC.

\begin{table*}[!htb]
\centering
\caption{Cross-domain transfer learning experiments performed on the Fault Diagnosis dataset. Domains A, B, C, and D represent the four working conditions. For supervised training, we train on the source domain and directly test on the target domain. For TS-TCC, we pretrain the model with the unlabeled source domain data, then fine-tune it with the few labels, then test it on the target domain. The same applies to CA-TCC, except that we include 1\% of labeled data in the training. Results are reported in terms of accuracy. \textbf{Best} results are in bold, while the \underline{second-best} ones are underlined.}
\begin{tabular}{@{}l|cccccccccccc|c@{}}
\toprule
Method & A$\rightarrow$B & A$\rightarrow$C & A$\rightarrow$D & B$\rightarrow$A & B$\rightarrow$C & B$\rightarrow$D & C$\rightarrow$A & C$\rightarrow$B & C$\rightarrow$D & D$\rightarrow$A & D$\rightarrow$B & D$\rightarrow$C & AVG \\ \midrule

Supervised & 34.38 & 44.94 & 34.57 & \textbf{52.93} & 63.67 & \underline{99.82} & \textbf{52.93} & 84.02 & 83.54 & \textbf{53.15} & 99.56 & 62.43 & 63.83 \\

TS-TCC & \underline{43.15} & \underline{51.50} & \underline{42.74} & \underline{47.98} & \underline{70.38} & 99.30 & 38.89 & \underline{98.31} & \underline{99.38} & \underline{51.91} & \underline{99.96} & \underline{70.31} & \underline{67.82} \\

CA-TCC & \textbf{44.75} & \textbf{52.09} & \textbf{45.63} & 46.26 & \textbf{71.33} & \textbf{100.0} & \underline{52.71} & \textbf{99.85} & \textbf{99.84} & 46.48 & \textbf{100.0} & \textbf{77.01} & \textbf{69.66}\\

\hline
\end{tabular}

\label{tbl:TL}
\end{table*}

\begin{table*}[!htb]
    \centering
    \caption{Results of different SEMI-SUPERVISED baselines with 1\% and 5\% of labels. \textbf{Best results} across each row are in bold, while the \underline{second-best results} are underlined.}
    \resizebox{\textwidth}{!}{
    \begin{tabular}{@{}l|cc|cc|cc|cc|cc|cc@{}}
    \toprule
    \multicolumn{13}{c}{\textbf{1\% of labeled data}}  \\ \midrule
     & \multicolumn{2}{c|}{Supervised} & \multicolumn{2}{c|}{Mean-Teacher} & \multicolumn{2}{c|}{DivideMix} & \multicolumn{2}{c|}{SemiTime} & \multicolumn{2}{c|}{FixMatch} &
     \multicolumn{2}{c}{CA-TCC} \\ \midrule
    Datasets & Accuracy & MF1-score & Accuracy & MF1-score & Accuracy & MF1-score & Accuracy & MF1-score & Accuracy & MF1-score & Accuracy & MF1-score \\ \midrule
            
            HAR & 44.9$\pm$6.7 & 41.0$\pm$6.8 & 75.9$\pm$1.9 & 74.0$\pm$2.8 & 76.5$\pm$0.7 & 75.4$\pm$1.0 & \textbf{77.6$\pm$1.1} & \textbf{76.3$\pm$0.9} & 76.4$\pm$2.5 & 75.6$\pm$2.8 & \underline{77.3$\pm$0.6} & \underline{76.2$\pm$0.1} \\
            
            Sleep-EDF & 34.1$\pm$0.3 & 30.8$\pm$0.8 & 73.6$\pm$1.0 & 63.7$\pm$0.3 & \underline{76.5$\pm$2.2} & \underline{66.6$\pm$0.8} & 73.6$\pm$3.9 & 63.4$\pm$2.9 & 72.5$\pm$2.2 & 62.0$\pm$2.9 & \textbf{79.4$\pm$0.1} & \textbf{70.8$\pm$0.5} \\
            
            Epilepsy & 76.1$\pm$0.7 & 74.8$\pm$0.4 & 91.5$\pm$0.3 & 90.6$\pm$0.6 & 90.9$\pm$0.7 & 89.4$\pm$1.4 & 91.6$\pm$0.3 & 90.8$\pm$0.6 & \textbf{93.2$\pm$0.2} & \textbf{92.2$\pm$0.5} & \underline{92.0$\pm$0.1} & \underline{91.9$\pm$0.1} \\
            
            
            Wafer & 91.9$\pm$1.3 & 67.6$\pm$9.2 & 94.7$\pm$0.2 & 84.7$\pm$0.3 & 93.2$\pm$0.5 & 82.0$\pm$0.8 & 94.4$\pm$0.6 & 84.4$\pm$1.2  & \underline{95.0$\pm$0.4} & \underline{84.8$\pm$1.2} & \textbf{95.1$\pm$0.3} & \textbf{85.1$\pm$0.6}  \\ 
            
            FordA & 56.4$\pm$1.6 & 54.4$\pm$3.5 & 71.7$\pm$1.6 & 71.5$\pm$1.8 & 73.7$\pm$1.1 & 73.3$\pm$0.9 & \underline{75.1$\pm$1.3} & \underline{74.4$\pm$1.4} & 74.5$\pm$0.4 & 74.3$\pm$0.4 & \textbf{82.3$\pm$1.1} & \textbf{81.7$\pm$1.3} \\ 
            
            FordB & 51.9$\pm$2.6 & 48.0$\pm$3.6 & 65.9$\pm$2.8 & 65.8$\pm$2.8 & 54.5$\pm$2.8 & 54.1$\pm$3.2 & \underline{67.6$\pm$2.2} & \underline{67.5$\pm$2.3} & 56.7$\pm$5.9 & 55.4$\pm$6.9 & \textbf{73.8$\pm$1.5} & \textbf{73.0$\pm$1.8}\\ 
            
            POC & 62.0$\pm$0.8 & 40.0$\pm$2.1 &  \underline{62.1$\pm$0.3} & \underline{40.8$\pm$1.2} & 62.1$\pm$0.6 & 40.7$\pm$2.1 & 62.0$\pm$0.5 & 40.4$\pm$1.6 & 61.9$\pm$0.5 & 40.0$\pm$1.8 & \textbf{63.4$\pm$0.4} & \textbf{49.3$\pm$0.7} \\ 
            
            PPOC & 64.3$\pm$0.7 & 64.2$\pm$0.7 & \textbf{65.3$\pm$1.9} & \textbf{64.6$\pm$1.3} & 56.1$\pm$6.9 & 55.6$\pm$7.2 & \underline{64.8$\pm$0.5} & \underline{64.6$\pm$0.6} & 63.7$\pm$1.9 & 63.5$\pm$1.8 & 63.4$\pm$0.3 & 63.1$\pm$0.3 \\ 
            
            StarLightCurves & 78.8$\pm$0.9 & 71.4$\pm$0.1 & 79.4$\pm$0.5 & 77.7$\pm$0.6 & 79.0$\pm$0.5 & 77.2$\pm$0.4 & \underline{79.5$\pm$0.5} & \underline{77.8$\pm$0.6} & 77.2$\pm$0.3 & 71.6$\pm$0.1 & \textbf{85.8$\pm$0.7} & \textbf{77.8$\pm$0.5} \\ 
            
            ElectricDevices & 57.8$\pm$1.1 & 47.5$\pm$1.0 & 48.9$\pm$8.2 & 48.3$\pm$1.5 & \underline{59.8$\pm$3.9} & \underline{49.4$\pm$2.6} & 57.3$\pm$3.7 & 48.1$\pm$2.7 & 58.2$\pm$1.0 & 46.9$\pm$0.4 & \textbf{65.9$\pm$0.8} & \textbf{56.7$\pm$1.1} \\

            \midrule
    
            Average & 61.8 & 54.0 & 72.9 & 68.2 & 72.2 & 66.4 & \underline{74.4} & \underline{68.8} & 72.9 & 66.6 & \textbf{77.8} & \textbf{72.6} \\
            
        \toprule
         \multicolumn{13}{c}{\textbf{5\% of labeled data}}  \\ 
        \toprule
        
        HAR & 52.8$\pm$1.5 & 50.9$\pm$0.2 & 88.2$\pm$1.2 & 88.1$\pm$1.2 & \textbf{89.1$\pm$2.0} & \textbf{89.1$\pm$1.3} & 87.6$\pm$1.3 & 87.1$\pm$0.8 & 87.6$\pm$0.3 & 87.3$\pm$0.4 & \underline{88.3$\pm$0.4} & \underline{88.3$\pm$0.3} \\
        
        Sleep-EDF & 60.5$\pm$3.9 & 54.8$\pm$5.5 & 75.2$\pm$0.4 & 64.8$\pm$0.8 & 75.4$\pm$1.3 & 65.4$\pm$1.4 & \underline{76.5$\pm$0.5} & \underline{65.9$\pm$0.9} & 75.7$\pm$1.5 & 65.1$\pm$2.2 & \textbf{82.1$\pm$0.2} & \textbf{74.6$\pm$0.1} \\
        
        Epilepsy & 83.4$\pm$0.7 &  80.4$\pm$0.6 & \underline{94.0$\pm$0.4} & \underline{93.6$\pm$0.7} & 93.9$\pm$0.6 & 93.4$\pm$1.1 & 94.0$\pm$0.5 & 93.0$\pm$0.9 & 93.7$\pm$1.4 & 92.4$\pm$0.3 & \textbf{94.5$\pm$0.1}  & \textbf{94.0$\pm$0.1} \\
        

        Wafer & 94.6$\pm$0.3 & 83.9$\pm$0.6 & 94.4$\pm$0.7 & 83.8$\pm$1.4 & 94.7$\pm$0.6 & 84.6$\pm$1.5 & \underline{95.0$\pm$0.4} & \underline{84.7$\pm$1.0} & 94.9$\pm$0.6 & 84.4$\pm$1.2 & \textbf{95.8$\pm$0.2} & \textbf{85.2$\pm$0.6} \\

        FordA & 54.5$\pm$4.3 & 44.1$\pm$8.0 & 82.6$\pm$1.6 & 82.5$\pm$1.7 & \underline{84.0$\pm$2.0} & \underline{83.9$\pm$2.1} & 83.8$\pm$1.5 & 83.7$\pm$1.5 & 83.8$\pm$2.2 & 83.8$\pm$2.3 & \textbf{90.9$\pm$0.3} & \textbf{90.8$\pm$0.3} \\ 
        
        FordB & 60.5$\pm$2.8 & 58.8$\pm$3.7 & 64.6$\pm$3.8 & 62.7$\pm$5.5 & 60.2$\pm$5.6 & 57.9$\pm$7.1 & \underline{65.0$\pm$4.9} & \underline{62.6$\pm$7.1} & 62.7$\pm$5.8 & 60.7$\pm$7.5 & \textbf{88.2$\pm$0.4} & \textbf{88.2$\pm$0.4} \\

        POC & 61.4$\pm$0.0 & 38.3$\pm$0.0 & 62.1$\pm$0.6 & 41.2$\pm$2.5 & 62.9$\pm$1.3 & 45.9$\pm$7.0 & 62.4$\pm$0.5 & 41.8$\pm$1.7 & \underline{63.1$\pm$1.4} & \underline{43.6$\pm$4.3} & \underline{66.4$\pm$0.3} & \underline{52.8$\pm$0.3} \\

        PPOC & 69.1$\pm$2.4 & 62.2$\pm$6.4 & \underline{73.4$\pm$7.6} & \underline{68.2$\pm$3.5} & 69.4$\pm$8.1 & 67.6$\pm$5.8 & 71.7$\pm$6.8 & 68.6$\pm$4.6 & 72.9$\pm$2.4 & 68.0$\pm$0.4 & \textbf{73.7$\pm$6.2} & \textbf{69.1$\pm$3.7} \\ 
        
        StarLightCurves & 81.8$\pm$0.8 & 71.4$\pm$4.1 & 84.9$\pm$2.0 & 83.9$\pm$1.4 & \underline{85.6$\pm$2.8} & \underline{84.1$\pm$2.1} & 84.6$\pm$4.8 & 83.8$\pm$3.7 & 84.1$\pm$2.0 & 77.5$\pm$3.0 & \textbf{88.8$\pm$0.7} & \textbf{81.1$\pm$2.0}\\ 
        
        ElectricDevices & 59.7$\pm$0.7 & 55.6$\pm$0.8 & 70.1$\pm$0.9 & 60.9$\pm$2.4 & \textbf{72.0$\pm$1.9} & \textbf{62.1$\pm$0.6} & \underline{71.6$\pm$1.0} & \underline{61.1$\pm$0.9} & 62.6$\pm$1.6 & 55.5$\pm$1.3 & 66.4$\pm$1.0 & 59.3$\pm$0.7 \\ 
        
        \midrule

        Average & 67.8 & 60.0 & 79.0 & 73.0 & 78.7 & \underline{73.4} & \underline{79.2} & 73.2 & 78.1 & 71.8 & \textbf{83.5} & \textbf{78.3} \\
    
    \bottomrule
    \end{tabular}
    }
    \label{tbl:ft_semi_ucr}
\end{table*}

\subsubsection{Performance under Different Labeling Budgets}
We investigate our TS-TCC performance under different fractions of few labels, by fine-tuning the pretrained model using  1\%, 5\%, 10\%, 50\%, and 75\% of randomly selected instances of the training data.
Fig.~\ref{Fig:few_labels} shows the results of our fine-tuned TS-TCC along with the supervised training under the aforementioned settings on three datasets. We evaluate the performance with MF1 because of the imbalance in the Sleep-EDF dataset.
In particular, TS-TCC fine-tuning (i.e., red curves in Fig.~\ref{Fig:few_labels}) means that we fine-tune the pretrained encoder with the few labeled samples. 

We observe that supervised training performs poorly with limited labeled data, while our TS-TCC fine-tuning achieves significantly better performance than supervised training. 
For example, with only 1\% of labeled data, TS-TCC fine-tuning can still achieve 69.5\% and 89.2\% for HAR and Epilepsy datasets respectively, compared with 47.6\% and 74.8\% for the supervised training on the same datasets. 
Furthermore, with only 10\% of labeled data, our fine-tuned TS-TCC can achieve comparable performance with the supervised training with 100\% of labeled data in the three datasets, demonstrating its effectiveness under different semi-supervised settings.

As shown in Fig.~\ref{Fig:few_labels}, TS-TCC fine-tuning is able to achieve impressive performance with few labeled data. Hence, TS-TCC is expected to generate high-quality pseudo labels in Fig.~\ref{Fig:supervised_contrastive}, which motivates us to propose the variant CA-TCC.

\subsection{Results of Semi-supervised CA-TCC Model}
\subsubsection{Comparison with Baseline Approaches}
To examine the efficacy of our CA-TCC in semi-supervised settings, we compare it against the following semi-supervised learning baselines.

\begin{enumerate}
    \item Mean-Teacher~\cite{mean_teacher}: it consists of two models. The first is the student model, which is the base regular model. The second is the teacher model, which is the averaging of the student model weights using the Exponential Moving Average (EMA) over the training steps.
    
    \item DivideMix~\cite{DivideMix}: it discards the labels that are likely to be noisy from the training and leverages them as unlabeled data to regularize the model against overfitting.

    \item SemiTime~\cite{semiTime}: it learns the temporal relations in unlabeled data by splitting the signal into past-future pairs, then contrasting the past of one sample with two future splits from another sample.
    
    \item FixMatch~\cite{sohn2020fixmatch}: it generates pseudo labels for the weak augmented view of the signal, and uses it to produce pseudo labels for the strong augmented view if it exceeds a confidence threshold.
    
\end{enumerate}
Notably, the four baselines train a cross-entropy loss on the labeled portion of the samples.
Also, for FixMatch, we apply our proposed weak and strong augmentations for training.

Table~\ref{tbl:ft_semi_ucr} compares the performance of CA-TCC against semi-supervised baselines across 10 different datasets. Overall, the average F1-score of our CA-TCC across the 10 datasets outperforms the average F1-score of the second-best performing semi-supervised baseline with 3.8\% and 4.9\% in 1\% and 5\% of labels, respectively. Specifically, our CA-TCC ranks first in seven datasets and second in two other datasets. The consistent superior performance of our CA-TCC across different datasets and labeling budgets demonstrates its effectiveness in maximizing the utilization of labeled data in semi-supervised settings.

\subsubsection{CA-TCC Model Analysis}
We discuss the different aspects that yield the improved performance of the CA-TCC.
Specifically, we regard the improved performance of CA-TCC to two main factors.
The first is the effective representations learned by TS-TCC in Phase~1, which enables generating high-quality pseudo labels. This is demonstrated in Table~\ref{tbl:ft_self_ucr}, where TS-TCC outperforms other self-supervised learning baselines. Therefore, it was anticipated that the fine-tuned TS-TCC encoder should be superior when used to produce the pseudo labels.
The second factor is the supervised contrastive loss in CA-TCC, which uses the pseudo-labeled data to include more positive pairs in the contrastive loss. 
Nevertheless, our CA-TCC is flexible in architecture and can anticipate different models throughout its different phases.
Next, we attempt to validate the aforementioned factors and support our conclusions.

\begin{figure}[!tbh]
    \centering
    \includegraphics[width=\columnwidth]{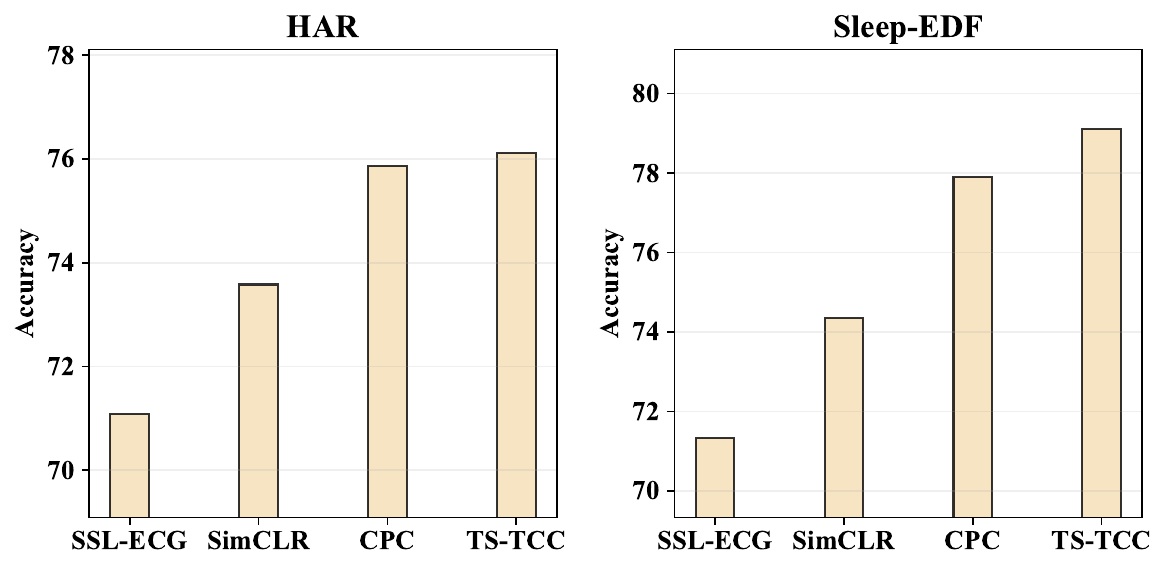}
    \caption{Accuracy of generated pseudo labels with different self-supervised learning methods applied to HAR and Sleep-EDF datasets. Fine-tuning in Phase 2 is performed with 1\% of labels.}
    \label{fig:pseudo_labels_acc}
\end{figure}

\paragraph{\textbf{Quality of pseudo labels}}
To investigate the effect of pseudo labels, we compare the quality of the pseudo labels generated by each baseline. Specifically, we replace our TS-TCC (in Phase~1) with SSL-ECG, SimCLR, and CPC. Then, for each baseline, we fine-tune the encoder and use it to generate pseudo labels (Phases~2 and~3) and then, conduct the semi-supervised training. Fig.~\ref{fig:pseudo_labels_acc} shows the accuracy of the generated pseudo labels (generated in Phase~3) compared to the true labels. We find that the pseudo labels generated by TS-TCC are more accurate than those generated by the other baselines. 
In addition, Table ~\ref{tbl:combinations} reports the performance with different baselines (deployed in Phase~1), while fixing CA-TCC (in Phase~4) for two datasets with different scales, i.e., HAR and Sleep-EDF. We observe that using TS-TCC in both phases results in the best performance, supporting our conclusion and justification for the first reason for improved performance.

\newcommand{\specialcell}[2][c]{%
  \begin{tabular}[#1]{@{}c@{}}#2\end{tabular}}

\begin{table}[!tbh]
\centering
\caption{Different combinations for the first and second pretraining phases in our framework. Pretraining in the first phase is fully unsupervised, while the second phase is class-aware (TS-TCC or SimCLR) with generated pseudo labels from fine-tuning with 1\% of labels.}
\resizebox{\columnwidth}{!}{
\begin{tabular}{@{}cc|cc|cc@{}}
\toprule
 & \multicolumn{1}{l|}{} & \multicolumn{2}{c|}{HAR} & \multicolumn{2}{c}{Sleep-EDF} \\ \midrule
\specialcell{Unsupervised \\ pretraining} & \specialcell{Semi-supervised \\ training} & Accuracy & MF1-score & \multicolumn{1}{c}{Accuracy} & \multicolumn{1}{c}{MF1-score} \\ \midrule

SSL-EEG & CA-(TS-TCC) & 70.8$\pm$0.9 & 68.4$\pm$0.8 & 76.1$\pm$0.4 & 67.1$\pm$0.6 \\
SimCLR & CA-(TS-TCC) & 73.2$\pm$1.2 & 70.5$\pm$1.2 & 78.3$\pm$0.7 & 68.1$\pm$0.5 \\
CPC & CA-(TS-TCC) & 72.7$\pm$0.7 & 69.6$\pm$0.8  & 79.2$\pm$0.4 & 70.0$\pm$0.6 \\
TS-TCC & CA-(TS-TCC) & \textbf{77.3$\pm$0.6} &  \textbf{76.2$\pm$0.1}  & \textbf{79.4$\pm$0.1} & \textbf{70.8$\pm$0.5} \\
\midrule
SSL-ECG & CA-(SimCLR) & 69.3$\pm$1.3 & 66.2$\pm$1.1 & 75.5$\pm$0.2 & 66.2$\pm$0.9 \\
SimCLR & CA-(SimCLR) & 67.5$\pm$1.4 & 64.3$\pm$1.2 & 77.5$\pm$0.2 & 67.9$\pm$0.6 \\
CPC & CA-(SimCLR) & 72.2$\pm$1.9 & 69.1$\pm$2.6 & 79.2$\pm$0.1 & 70.1$\pm$0.7  \\
TS-TCC & CA-(SimCLR) & \underline{75.0$\pm$1.5} & \underline{73.1$\pm$2.0}  & \underline{79.3$\pm$0.3} & \underline{70.2$\pm$0.1} \\ \bottomrule
\end{tabular}
}
\label{tbl:combinations}
\end{table}

\paragraph{\textbf{Effect of supervised contrastive loss}}
In this experiment, we compare the performance of the semi-supervised training (Phase~4) when using the supervised contrastive loss against the unsupervised contrastive loss.
The supervised contrastive loss considers samples having the same class label as positive pairs and samples from different classes as negative pairs. In contrast, the unsupervised contrastive loss only forms positive pairs from the augmented views of the sample, and all other samples in the mini-batch are considered as negative pairs. This difference can affect the performance of the model, as the unsupervised contrastive loss may treat samples having a similar class as negative pairs.

\begin{table*}[!htb]
\centering
\caption{Ablation study of the effect of different components in TS-TCC and CA-TCC models. We also show the effect of using two weak or two strong augmentations on their performance. It is clear that using a combination of weak and strong augmentations yields the best performance. The results are obtained with the \textit{linear evaluation} experiment on 5\% of labeled data on three datasets.}
\begin{tabular}{@{}l|cc|cc|cc@{}}
\toprule
 & \multicolumn{2}{c|}{HAR} & \multicolumn{2}{c|}{Sleep-EDF} & \multicolumn{2}{c}{Epilepsy} \\ \midrule
Component & Accuracy & MF1-score & Accuracy & MF1-score & Accuracy & MF1-score \\ \midrule

TC only  & 68.16$\pm$1.15 & 66.89$\pm$1.11 & 75.55$\pm$0.93 & 60.19$\pm$0.81 & 88.29$\pm$1.29 & 88.00$\pm$1.91 \\

TC + X-Aug & 74.22$\pm$1.03 & 72.18$\pm$0.99 & 77.80$\pm$0.29 & 61.28$\pm$1.22 & 90.51$\pm$0.43 & 89.27$\pm$0.22 \\ 

TS-TCC (TC + X-Aug + CC) & 77.58$\pm$1.78 & 76.66$\pm$1.96 & 76.98$\pm$0.56 & 70.94$\pm$0.46 & 93.12$\pm$0.31 & 93.67$\pm$0.56 \\ 

CA-TCC (TC + X-Aug + SCC)  & \textbf{88.27$\pm$0.38} & \textbf{88.29$\pm$0.34} & \textbf{82.14$\pm$0.19} & \textbf{74.75$\pm$0.06} & \textbf{94.52$\pm$0.12} & \textbf{94.00$\pm$0.09} \\   
\midrule

TS-TCC (Weak only)   & 67.39$\pm$1.73 & 65.54$\pm$2.42 & 79.63$\pm$0.16 & 68.15$\pm$0.23 & 93.22$\pm$0.11 & 91.97$\pm$0.19 \\ 
CA-TCC (Weak only)   & 85.68$\pm$0.26 & 84.77$\pm$0.25 & 81.62$\pm$0.89 & 70.10$\pm$1.28 & 93.84$\pm$0.05 & 92.19$\pm$0.10 \\ 

\midrule

TS-TCC (Strong only)  & 50.37$\pm$1.18 & 43.05$\pm$1.42 & 74.84$\pm$0.50 & 64.53$\pm$0.58 & 92.49$\pm$0.62 & 90.60$\pm$0.20 \\ 
CA-TCC (Strong only)  & 59.59$\pm$0.06 & 53.34$\pm$0.49 & 79.24$\pm$0.74 & 69.39$\pm$0.89 & 93.74$\pm$0.04 & 92.00$\pm$0.05 \\

\bottomrule
\end{tabular}

\label{tbl:ablation}
\end{table*}

\begin{figure}[!tb]
    \centering
    \includegraphics[width=\columnwidth]{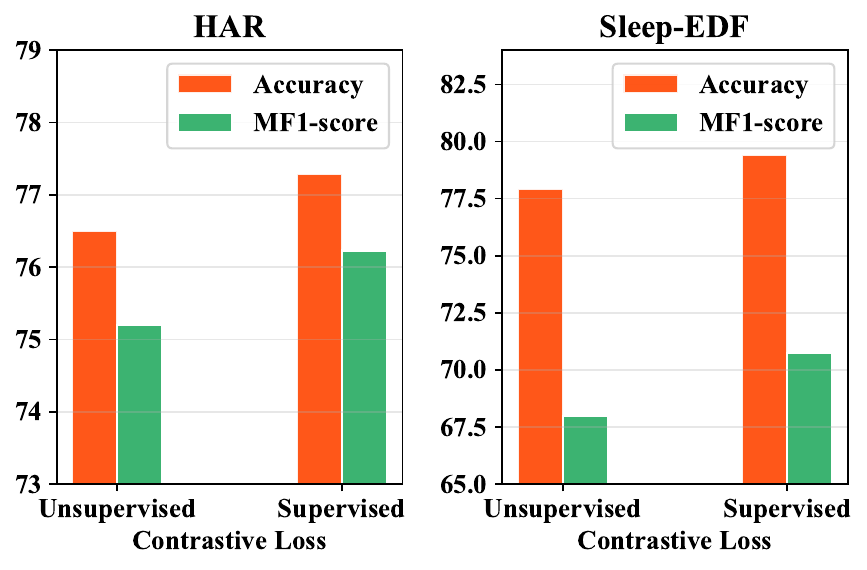}
    \caption{Performance comparison with and without supervised contrastive loss in the semi-supervised training (Phase~4).}
    \label{fig:comp_unsup_sup_cont_loss}
\end{figure}

Fig.~\ref{fig:comp_unsup_sup_cont_loss} compares the performance of the unsupervised and the supervised contrastive losses in terms of accuracy and F1-score on the HAR and Sleep-EDF datasets. The results indicate that the use of the supervised contrastive loss, which includes more positive pairs, results in improved performance and better representation learning on both datasets.

\paragraph{\textbf{Varying CA-TCC model architecture}}
In this section, we evaluate the performance of different combinations of self-supervised learning algorithms in Phases~1 and 4. Specifically, we use different self-supervised algorithms in Phase~1, while in Phase~4, we deploy only two baselines that can use the supervised contrastive loss, i.e., SimCLR and TS-TCC. The experiments are conducted on the HAR and Sleep-EDF datasets using 1\% of labels.

The results, shown in Table~\ref{tbl:combinations}, indicate that using class-aware TS-TCC in the semi-supervised training with any baseline in Phase~1 consistently outperforms class-aware SimCLR for both datasets. This superior performance of TS-TCC can be regarded to its ability to learn temporal relations in time-series data while also benefiting from the supervised contrasting, unlike class-aware SimCLR which only learns through contrasting positive and negative pairs. 

In summary, our framework is flexible and allows using various models in different training phases. However, we chose to only use TS-TCC as it is the best-performing self-supervised learning model among the baselines and it is able to handle class-aware training in semi-supervised settings.

\begin{figure*}
     \centering
     \begin{subfigure}[b]{0.3\textwidth}
         \centering
         \includegraphics[width=\textwidth]{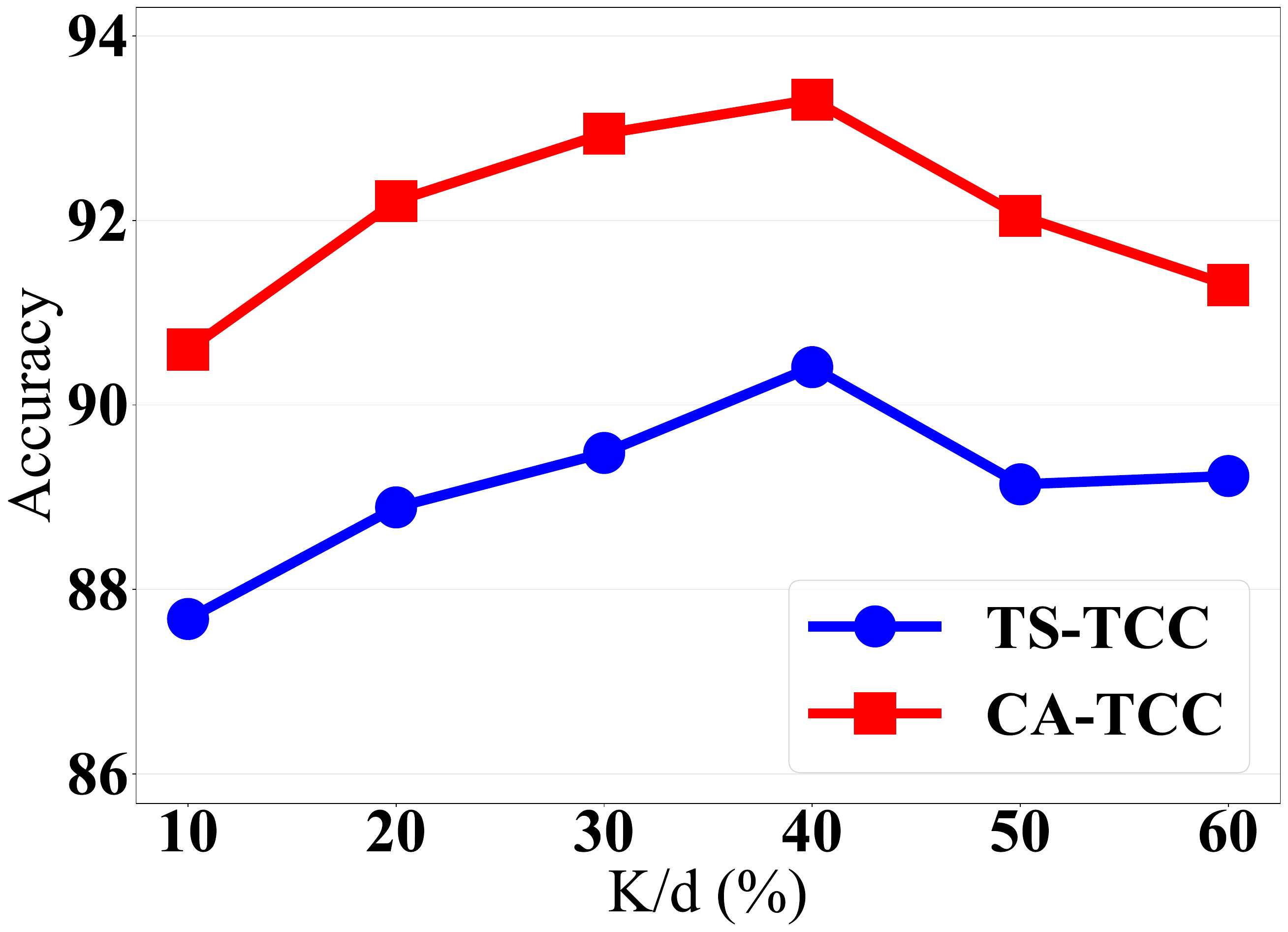}
         \caption{}
         \label{fig:sens1}
     \end{subfigure}
     \hfill
     \begin{subfigure}[b]{0.3\textwidth}
         \centering
         \includegraphics[width=\textwidth]{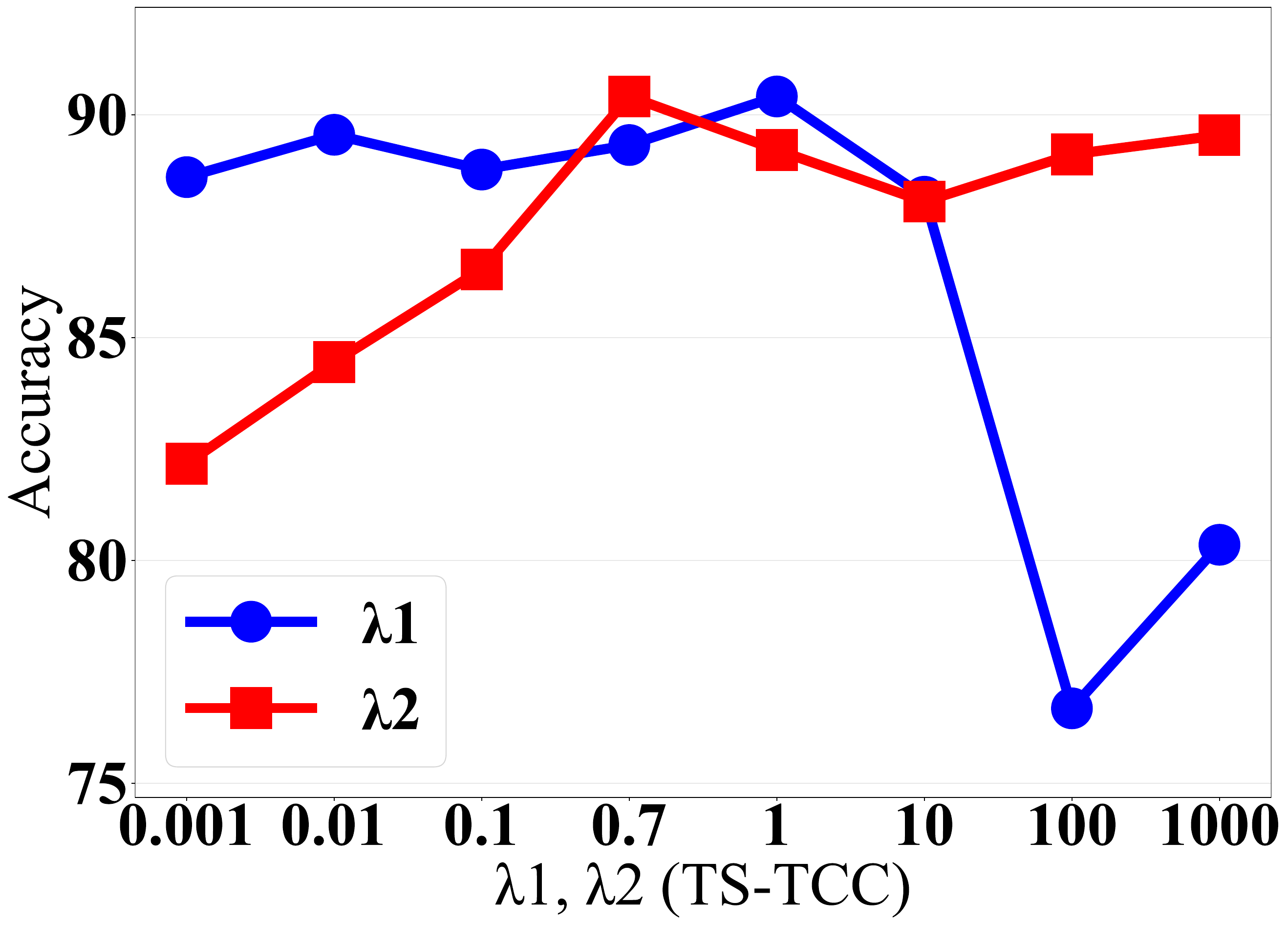}
         \caption{}
         \label{fig:sens2}
     \end{subfigure}
     \hfill
     \begin{subfigure}[b]{0.3\textwidth}
         \centering
         \includegraphics[width=\textwidth]{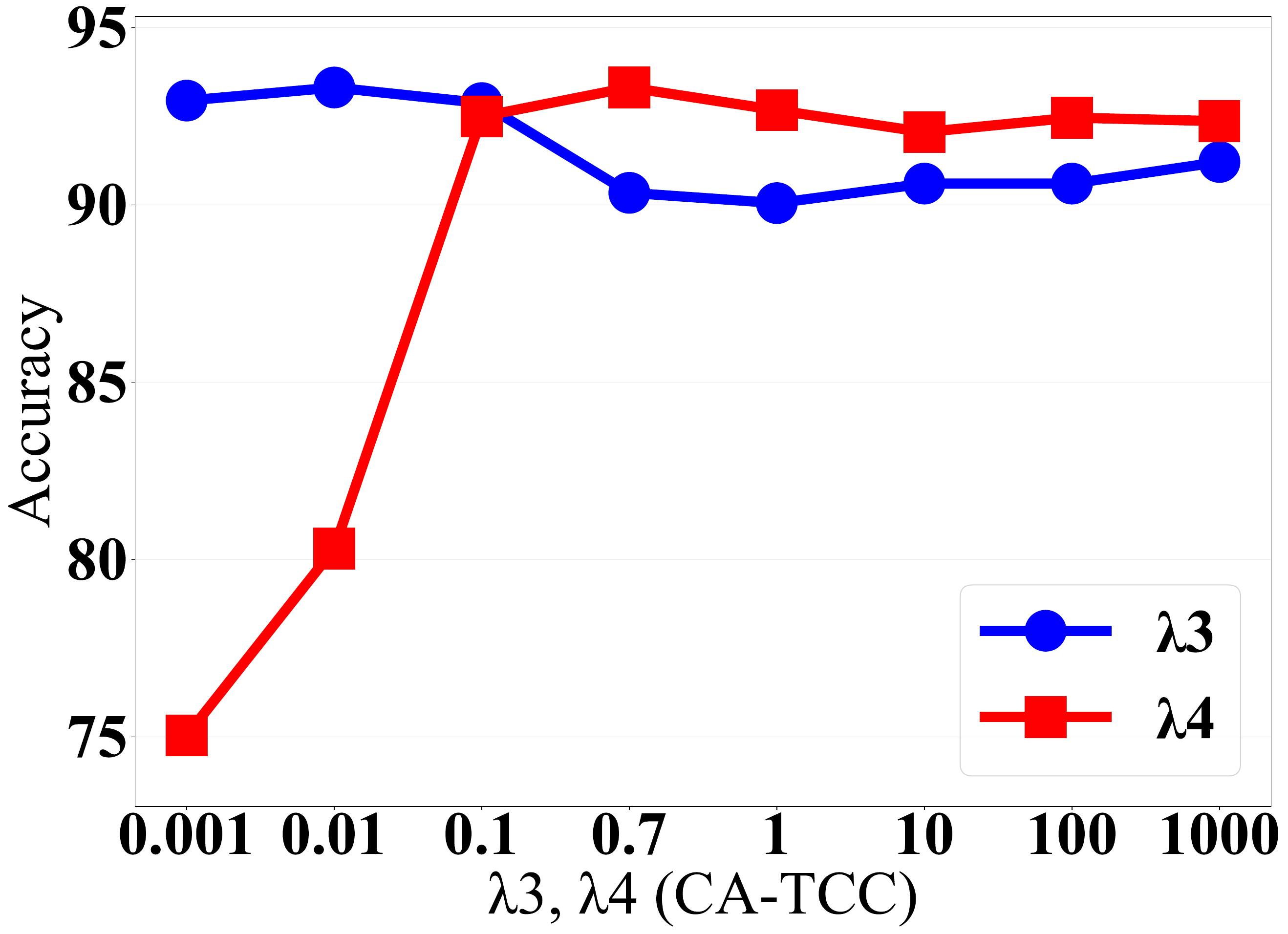}
         \caption{}
         \label{fig:sens3}
     \end{subfigure}
        \caption{Sensitivity analysis experiments on HAR dataset. Figure (a) shows the effect of changing the percentage of the predicted future timesteps, where we notice a close performance among our two variants. Figure (b) shows the impact of the different combinations of $\lambda_1$ and $\lambda_2$ on TS-TCC performance. Last, figure (c) shows the effect of the different combinations of $\lambda_3$ and $\lambda_4$ onn CA-TCC performance.}
        \label{Fig:ts_sens_analysis}
\end{figure*}

\subsection{Transfer Learning Experiment}
We further examine the transferability of the learned features by designing a transfer learning experiment. We use Fault Diagnosis (FD) dataset for the evaluation under the transfer learning setting. Recall that the FD dataset has four working conditions, which are considered as four domains (denoted as domains A, B, C, and D). Here, we train the model on the data from one condition (i.e., source domain) and test it on another condition (i.e., target domain). We adopt three training schemes on the source domain, namely, (1) Supervised training, (2) TS-TCC fine-tuning, and (3) CA-TCC fine-tuning. In TS-TCC and CA-TCC fine-tuning, we fine-tune our pretrained encoder using the labeled data in the source domain.

Table~\ref{tbl:TL} shows the performance of the three training schemes under 12 cross-domain scenarios. Clearly, our pretrained TS-TCC model consistently outperforms the supervised pretraining in 8 out of 12 cross-domain scenarios. Similarly, with only 1\% of labels in each source domain for training, we find that CA-TCC model outperforms the supervised pretraining in 9 out of 12 cross-domain scenarios. We find that TS-TCC model can achieve at least $\sim$7\% improvement in 7 out of 8 winning scenarios (except for D$\rightarrow$B scenario). Similarly, CA-TCC model can achieve at least $\sim$8\% improvement in 7 out of 9 winning scenarios. Overall, our two proposed approaches can improve the transferability of learned representations over the supervised training by $\sim$ 4\% and 6\% in terms of accuracy.

\subsection{Ablation Study}
We study the effectiveness of each component in our proposed CA-TCC model.
Specifically, we derive different model variants for comparison as follows. First, we train the Temporal Contrasting (TC) module without the cross-view prediction task, where each branch predicts the future timesteps of the same augmented view. This variant is denoted as `TC only'. Second, we train the TC module with adding the cross-view prediction task, which is denoted as `TC + X-Aug'. Third, we train the proposed TS-TCC model, which is denoted as `TC + X-Aug + CC'. Finally, we train the proposed CA-TCC model, which is denoted as `TC + X-Aug + SCC'. We also study the effect of using a single family of augmentations on the performance of TS-TCC and CA-TCC. In particular, for an input $x$, we generate two different views $x_1$ and $x_2$ from the same augmentation type, i.e., $x_1 \!\sim\! \mathcal{T}_w$ and $x_2 \!\sim\! \mathcal{T}_w$ when using either the weak augmentation or the strong augmentation. We show the linear evaluation results in terms of accuracy and macro F1-score with only 5\% throughout these experiments.

Table~\ref{tbl:ablation} shows this ablation study on the three datasets.
Clearly, the proposed cross-view prediction task generates robust features and thus improves the performance by more than 6\% accuracy on HAR datasets, and $\sim$2\% on Sleep-EDF and Epilepsy datasets. Additionally, the contextual contrasting module further improves the performance, as it helps the features to be more discriminative. 
More improvement was achieved by using supervised contextual contrasting in CA-TCC, which supports the importance of considering more positive samples from the same class to generate more discriminative features.
By studying the effect of augmentations on TS-TCC, we find that generating different views from the same augmentation type is not helpful with HAR and Sleep-EDF datasets. For these complex datasets, using only weak augmentations may not make a \textit{tough} cross-view prediction task, leading to close results to the variant `TC only'. Counterpart, using only strong augmentations deviates the model from recognizing the original data while testing.
However, the less complex Epilepsy dataset can still achieve comparable performance with only one augmentation. For CA-TCC, we find that it consistently outperforms the results of TS-TCC, showing its effectiveness to improve the representations with the available few labeled samples. For example, we find that it highly improves the performance of using only weak or strong augmentations in both HAR and Sleep-EDF datasets.

\begin{figure*}
     \centering
     \begin{subfigure}[b]{0.3\textwidth}
         \centering
         \includegraphics[width=\textwidth]{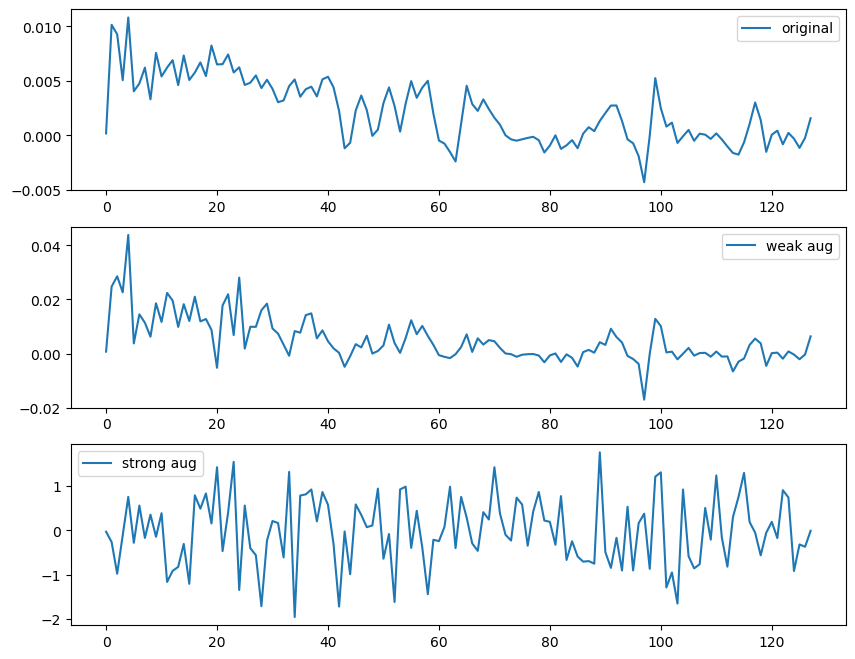}
         \caption{HAR}
         \label{fig:har_aug}
     \end{subfigure}
     \hfill
     \begin{subfigure}[b]{0.3\textwidth}
         \centering
         \includegraphics[width=\textwidth]{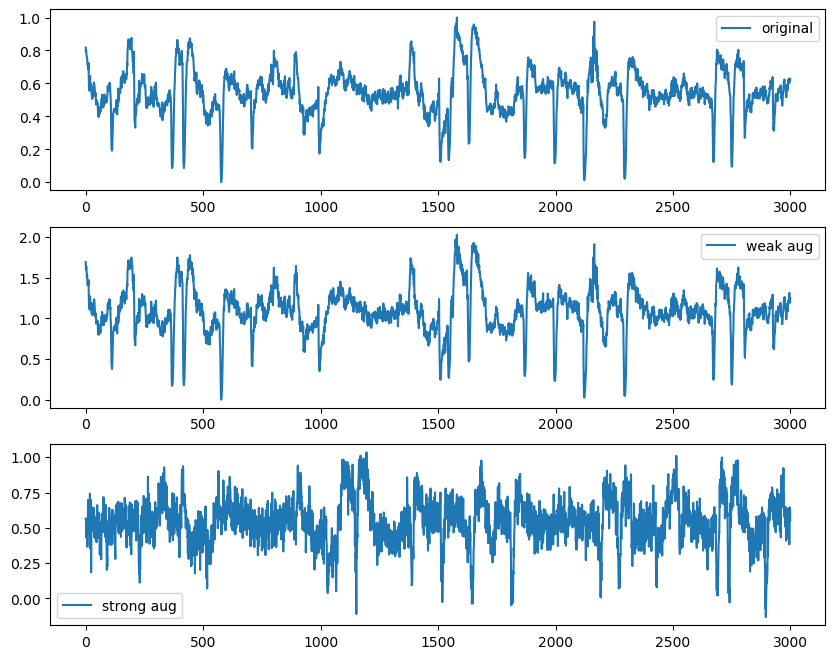}
         \caption{Sleep-EDF}
         \label{fig:eeg_aug}
     \end{subfigure}
     \hfill
     \begin{subfigure}[b]{0.3\textwidth}
         \centering
         \includegraphics[width=\textwidth]{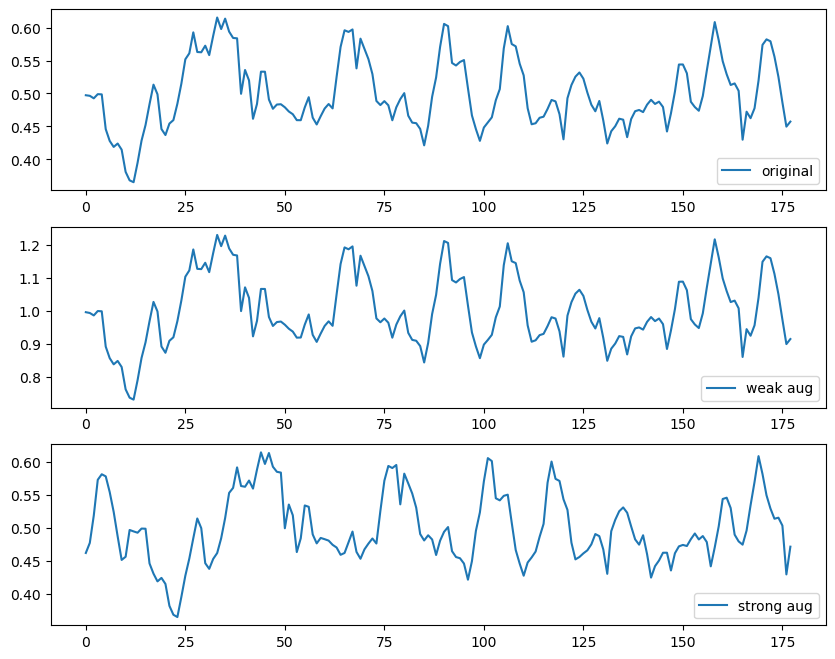}
         \caption{Epilepsy}
         \label{fig:epilepsy_aug}
     \end{subfigure}
        \caption{Sample from each adopted dataset after normalization along with its weak and strong augmented views. The first row is the original samples, the second row shows the weak augmented views, and the third row shows the strong augmented views.}
        \label{Fig:ts_data}
\end{figure*}

\subsection{Sensitivity Analysis}
\label{sec:sens_analysis}
We perform sensitivity analysis on the HAR dataset to study five parameters namely, the number of predicted future timesteps $K$ in the temporal contrasting module, $\lambda_1$ and $\lambda_2$ in Eq.~\ref{eq:overall}, and $\lambda_3$ and $\lambda_4$ in Eq.~\ref{eq:overall_2}. In specific, we used the linear evaluation experiment with full labels to assess the performance.

Fig.~\ref{fig:sens1} shows the effect of $K$ on the overall performance of TS-TCC and CA-TCC, where the x-axis is the percentage $K/d$, and $d$ is the length of the features. Clearly, increasing the percentage of the predicted future timesteps improves the performance. However, larger percentages can harm the performance as it reduces the amount of past data used for training the autoregressive model. We observe that predicting 40\% of the total feature length performs the best, and thus we set $K$ as $d\times$40\% in our experiments. The same conclusion can be drawn for both variants of our framework. Fig.~\ref{fig:sens2} shows the results of varying $\lambda_1$ and $\lambda_2$ in TS-TCC (Eq.~\ref{eq:overall}) in a range between 0.001 and 1000 respectively. We first fix $\lambda_1=1$ and change the values of $\lambda_2$. We observe that our model achieves good performance when $\lambda_2 \approx 1$, where the model performs best with $\lambda_2=0.7$. 
Consequently, we fix $\lambda_2=0.7$ and tune the value of $\lambda_1$ as in Fig.~\ref{fig:sens2}, where we find that our model achieves the best performance when $\lambda_1=1$. We also find that as $\lambda_1<10$, our model is less sensitive to its value, while it is more sensitive to different values of $\lambda_2$.

We also perform sensitivity analysis on the values of $\lambda_3$ and $\lambda_4$ in CA-TCC (Eq.~\ref{eq:overall_2}) in a similar manner and within the same ranges, as shown in Fig.~\ref{fig:sens3}. We also used 1\% of labels to generate pseudo labels. Consequently, we chose the values of $\lambda_3=0.01$ and $\lambda_4=0.7$.

\subsection{Augmentation Selection}
\label{sec:aug_selec}

The proper selection of augmentations is crucial for contrastive learning techniques, as contrastive methods are sensitive to the choice of augmentations \cite{simclr_paper}. Many studies showed that composing multiple data augmentation operations achieves better performance for image data \cite{simclr_paper,khosla2020supervised}.
However, the selection of the proper augmentations that well-fit time-series data is less explored and still an open problem \cite{wen2021time}. Here, we aim to study the choice of suitable augmentations for our contrastive learning problem.

\begin{table}[!tbh]
\centering
\caption{A study of TS-TCC \textit{linear evaluation} performance with 5\% of labeled HAR data when using different variations of weak and strong augmentations.}
\begin{tabular}{@{}l|l|cc@{}}
\toprule
Weak Augmentation    & Strong Augmentation   & Accuracy  & MF1-score  \\ \midrule
Scale                & No Augmentation                & 46.12 & 36.67 \\
Scale + Jitter       & No Augmentation                & 56.98 & 50.88 \\

No Augmentation                & Permutation          & 62.07 & 52.64 \\
No Augmentation                & Jitter + Permutation & 72.35 & 68.03 \\

Time Shift            & Jitter + Permutation & 71.57 & 66.97 \\
Time Shift + Jitter   & Jitter + Permutation & \underline{74.69} & \underline{69.33} \\

Scale                 & Jitter + Permutation & 72.59 & 68.90 \\
Scale + Jitter       & Jitter + Permutation  & \textbf{77.58} & \textbf{76.66} \\

\bottomrule
\end{tabular}
\label{tbl:augmentations_study}
\end{table}

We define the weak augmentation as the one that applies limited change on the shape of the original signal as shown in the second row of Fig.~\ref{Fig:ts_data}. Examples of weak augmentations include scaling and time shifting. On the other hand, strong augmentation makes strong perturbations on the signal shape with keeping some of its temporal information, such as permutation. This is shown in the third row of Fig.~\ref{Fig:ts_data}. Permutation includes splitting the signal into $M$ chunks and shuffling their order. Notably, some augmentations such as jittering (i.e., adding random noise) can be considered as both strong and weak augmentations depending on the added noise level. To justify the selection of our proposed augmentations, we provide a systematic study on the impact of applying several different data augmentations for each view in TS-TCC (Fig.~\ref{Fig:overall}). The results are provided in Table~\ref{tbl:augmentations_study}.

We first apply weak augmentation only, i.e., scaling, to one view without applying any augmentation to the other view, and the accuracy is only 46.12\% as shown in Table~\ref{tbl:augmentations_study}. However, by composing jitter to scaling, we can observe a large performance improvement. Meanwhile, we also apply strong augmentation only and keep the other view without augmentations. The accuracy, in this case, is 62.07\%, which is much higher than applying only weak augmentation. Besides, we notice that adding a jitter to the permutation can further improve the accuracy to 72.35\%.

We also test to apply another weak augmentation that does not highly affect the signal characteristics along with the strong augmentation. We can find that using time shift as a weak augmentation achieves relatively good performance, i.e., an accuracy of 71.57\%. Again, it can be found that adding jitter to time shift further improves the accuracy to 74.69\%. Similar results are achieved when replacing time-shift with scaling.
In addition, applying only weak or strong augmentations to both views results in poorer performance as shown in Table~\ref{tbl:ablation}. 
At this point, we find that applying weak augmentation for one view and strong augmentations for other views achieves the best performance, i.e., an accuracy of 77.58\%.
As an explanation, weak augmentations apply limited changes to the shape of the original signal as shown in Fig.~\ref{Fig:ts_data}, which helps the model to perform well on the test data.

Via observing the different characteristics of three different samples in Fig. \ref{Fig:ts_data} in terms of signal magnitude and sampling rates, we conclude that the choice of proper parameters for augmentations (e.g., jitter and scaling ratio) will vary from one dataset to another.
Consequently, the range of parameter choices will be highly dependent on the characteristics of each time-series data.
Therefore, we propose to normalize the signals as a preprocessing step in our framework to improve parameter selection.
For example, the value of added jitter after normalizing the signals in weak augmentation should be less than the ones added in the strong augmentation.
In the experiments, we observe that the best practice is to normalize the data between 0 and 1 and set the weak jitter to be in the range [0, 0.1] and the strong jitter in the range [0.1, 1].
Similarly, a scaling ratio of 2 would be sufficient for weak augmentation in any time-series signals.

Similarly, for strong augmentation, it is important to properly select the number of chunks $M$, where the value of $M$ in time-series data with longer sequences should be greater than its value in those with shorter sequences. 
We find that selecting 40\% of the total feature size achieves the best performance for the three time-series datasets (see Section \ref{sec:sens_analysis}).

\section{Conclusion}
In this paper, we proposed two representation learning frameworks for time-series data. The first, i.e., TS-TCC, is for self-supervised learning from unlabeled data, while the second, i.e., CA-TCC is for semi-supervised learning when few labeled samples are available. In particular, we propose time-series-specific weak and strong augmentations and provide a systematic study of the choice of these augmentations. We use these augmentations to learn transformation-invariant representations through our proposed temporal and contextual contrasting modules in TS-TCC. By training a linear classifier with few labels on top of the learned representation by TS-TCC, it achieved comparable performance to the fully-supervised training. In addition, TS-TCC showed noticeable improvement under different labeling budgets, where 10\% of the labeled data could achieve close performance to the supervised training with full labeled data. We extended TS-TCC to the semi-supervised settings and proposed CA-TCC, which benefits from the pseudo labels generated by the fine-tuned TS-TCC model to train a class-aware supervised contrastive loss. CA-TCC was able to further improve this performance with only 1\% of labeled data when testing on different datasets. In addition, both variants were able to improve the transferability of the representations in real-world transfer learning scenarios.

\bibliographystyle{unsrt}
\bibliography{citations}

\begin{thebibliography}{10}

\bibitem{gharehbaghi2017deep}
Arash Gharehbaghi and Maria Lind{\'e}n.
\newblock A deep machine learning method for classifying cyclic time series of
  biological signals using time-growing neural network.
\newblock {\em IEEE transactions on neural networks and learning systems},
  29(9):4102--4115, 2017.

\bibitem{ching2018opportunities}
Travers Ching, Daniel~S Himmelstein, Brett~K Beaulieu-Jones, Alexandr~A
  Kalinin, Brian~T Do, Gregory~P Way, Enrico Ferrero, Paul-Michael Agapow,
  Michael Zietz, Michael~M Hoffman, et~al.
\newblock Opportunities and obstacles for deep learning in biology and
  medicine.
\newblock {\em Journal of The Royal Society Interface}, 2018.

\bibitem{puzzle}
Mehdi Noroozi and Paolo Favaro.
\newblock Unsupervised learning of visual representations by solving jigsaw
  puzzles.
\newblock In Bastian Leibe, Jiri Matas, Nicu Sebe, and Max Welling, editors,
  {\em Computer Vision -- ECCV}, pages 69--84, Cham, 2016. Springer
  International Publishing.

\bibitem{simclr_paper}
Ting Chen, Simon Kornblith, Mohammad Norouzi, and Geoffrey Hinton.
\newblock A simple framework for contrastive learning of visual
  representations.
\newblock In {\em Proceedings of the 37th International Conference on Machine
  Learning}, ICML'20. JMLR.org, 2020.

\bibitem{gidaris_unsupervised}
Spyros Gidaris, Praveer Singh, and Nikos Komodakis.
\newblock Unsupervised representation learning by predicting image rotations.
\newblock In {\em International Conference on Learning Representations}, 2018.

\bibitem{hjelm2018learning}
R~Devon Hjelm, Alex Fedorov, Samuel Lavoie-Marchildon, Karan Grewal, Phil
  Bachman, Adam Trischler, and Yoshua Bengio.
\newblock Learning deep representations by mutual information estimation and
  maximization.
\newblock In {\em International Conference on Learning Representations}, 2019.

\bibitem{He_2020_CVPR}
Kaiming He, Haoqi Fan, Yuxin Wu, Saining Xie, and Ross Girshick.
\newblock Momentum contrast for unsupervised visual representation learning.
\newblock In {\em Proceedings of the IEEE/CVF Conference on Computer Vision and
  Pattern Recognition (CVPR)}, June 2020.

\bibitem{NEURIPS2019_53c6de78}
Jean-Yves Franceschi, Aymeric Dieuleveut, and Martin Jaggi.
\newblock Unsupervised scalable representation learning for multivariate time
  series.
\newblock In H.~Wallach, H.~Larochelle, A.~Beygelzimer, F.~d\textquotesingle
  Alch\'{e}-Buc, E.~Fox, and R.~Garnett, editors, {\em Advances in Neural
  Information Processing Systems}, volume~32. Curran Associates, Inc., 2019.

\bibitem{pmlr-v136-mohsenvand20a}
Mostafa~Neo Mohsenvand, Mohammad~Rasool Izadi, and Pattie Maes.
\newblock Contrastive representation learning for electroencephalogram
  classification.
\newblock In Emily Alsentzer, Matthew B.~A. McDermott, Fabian Falck,
  Suproteem~K. Sarkar, Subhrajit Roy, and Stephanie~L. Hyland, editors, {\em
  Proceedings of the Machine Learning for Health NeurIPS Workshop}, volume 136
  of {\em Proceedings of Machine Learning Research}, pages 238--253. PMLR, 11
  Dec 2020.

\bibitem{cheng2020subject}
Joseph~Y Cheng, Hanlin Goh, Kaan Dogrusoz, Oncel Tuzel, and Erdrin Azemi.
\newblock Subject-aware contrastive learning for biosignals.
\newblock {\em arXiv preprint arXiv:2007.04871}, 2020.

\bibitem{eldele_ts_tcc}
Emadeldeen Eldele, Mohamed Ragab, Zhenghua Chen, Min Wu, Chee~Keong Kwoh,
  Xiaoli Li, and Cuntai Guan.
\newblock Time-series representation learning via temporal and contextual
  contrasting.
\newblock In Zhi-Hua Zhou, editor, {\em Proceedings of the Thirtieth
  International Joint Conference on Artificial Intelligence, {IJCAI-21}}, pages
  2352--2359. International Joint Conferences on Artificial Intelligence
  Organization, 8 2021.
\newblock Main Track.

\bibitem{qi2020small}
Guo-Jun Qi and Jiebo Luo.
\newblock Small data challenges in big data era: A survey of recent progress on
  unsupervised and semi-supervised methods.
\newblock {\em IEEE Transactions on Pattern Analysis and Machine Intelligence},
  44(4):2168--2187, 2022.

\bibitem{jing2020self}
Longlong Jing and Yingli Tian.
\newblock Self-supervised visual feature learning with deep neural networks: A
  survey.
\newblock {\em IEEE Transactions on Pattern Analysis and Machine Intelligence},
  43(11):4037--4058, 2021.

\bibitem{jaiswal2020a}
Ashish Jaiswal, Ashwin~Ramesh Babu, Mohammad~Zaki Zadeh, Debapriya Banerjee,
  and Fillia Makedon.
\newblock A survey on contrastive self-supervised learning.
\newblock {\em Technologies}, 9(1), 2021.

\bibitem{zhang2016colorful}
Richard Zhang, Phillip Isola, and Alexei~A. Efros.
\newblock Colorful image colorization.
\newblock In Bastian Leibe, Jiri Matas, Nicu Sebe, and Max Welling, editors,
  {\em Computer Vision -- ECCV}, pages 649--666, Cham, 2016. Springer
  International Publishing.

\bibitem{misra2020self}
Ishan Misra and Laurens van~der Maaten.
\newblock Self-supervised learning of pretext-invariant representations.
\newblock In {\em Proceedings of the IEEE/CVF Conference on Computer Vision and
  Pattern Recognition (CVPR)}, June 2020.

\bibitem{Zhai_2019_ICCV}
Xiaohua Zhai, Avital Oliver, Alexander Kolesnikov, and Lucas Beyer.
\newblock S4l: Self-supervised semi-supervised learning.
\newblock In {\em IEEE/CVF International Conference on Computer Vision (ICCV)},
  pages 1476--1485, 2019.

\bibitem{srivastava2015unsupervised}
Nitish Srivastava, Elman Mansimov, and Ruslan Salakhutdinov.
\newblock Unsupervised learning of video representations using lstms.
\newblock In {\em Proceedings of the 32nd International Conference on
  International Conference on Machine Learning - Volume 37}, ICML'15, page
  843–852. JMLR.org, 2015.

\bibitem{wei2018learning}
Donglai Wei, Joseph~J. Lim, Andrew Zisserman, and William~T. Freeman.
\newblock Learning and using the arrow of time.
\newblock In {\em Proceedings of the IEEE Conference on Computer Vision and
  Pattern Recognition (CVPR)}, June 2018.

\bibitem{liu2019exploiting}
Xialei Liu, Joost van~de Weijer, and Andrew~D. Bagdanov.
\newblock Exploiting unlabeled data in cnns by self-supervised learning to
  rank.
\newblock {\em IEEE Transactions on Pattern Analysis and Machine Intelligence},
  41(8):1862--1878, 2019.

\bibitem{oord2018representation}
Aaron van~den {Oord}, Yazhe {Li}, and Oriol {Vinyals}.
\newblock Representation learning with contrastive predictive coding.
\newblock {\em arXiv: Learning}, 2018.

\bibitem{grill2020bootstrap}
Jean-Bastien Grill, Florian Strub, Florent Altch\'{e}, Corentin Tallec,
  Pierre~H. Richemond, Elena Buchatskaya, Carl Doersch, Bernardo~Avila Pires,
  Zhaohan~Daniel Guo, Mohammad~Gheshlaghi Azar, Bilal Piot, Koray Kavukcuoglu,
  R\'{e}mi Munos, and Michal Valko.
\newblock Bootstrap your own latent a new approach to self-supervised learning.
\newblock In {\em Proceedings of the 34th International Conference on Neural
  Information Processing Systems}, NIPS'20, Red Hook, NY, USA, 2020. Curran
  Associates Inc.

\bibitem{simsiam}
Xinlei Chen and Kaiming He.
\newblock Exploring simple siamese representation learning.
\newblock In {\em Proceedings of the IEEE/CVF Conference on Computer Vision and
  Pattern Recognition (CVPR)}, pages 15750--15758, June 2021.

\bibitem{sohn2020fixmatch}
Kihyuk Sohn, David Berthelot, Nicholas Carlini, Zizhao Zhang, Han Zhang,
  Colin~A Raffel, Ekin~Dogus Cubuk, Alexey Kurakin, and Chun-Liang Li.
\newblock Fixmatch: Simplifying semi-supervised learning with consistency and
  confidence.
\newblock In H.~Larochelle, M.~Ranzato, R.~Hadsell, M.F. Balcan, and H.~Lin,
  editors, {\em Advances in Neural Information Processing Systems}, volume~33,
  pages 596--608. Curran Associates, Inc., 2020.

\bibitem{SSL_har}
Aaqib Saeed, Tanir Ozcelebi, and Johan Lukkien.
\newblock Multi-task self-supervised learning for human activity detection.
\newblock {\em Proc. ACM Interact. Mob. Wearable Ubiquitous Technol.}, 3(2),
  jun 2019.

\bibitem{ecg_emotion_rec}
Pritam Sarkar and Ali Etemad.
\newblock Self-supervised ecg representation learning for emotion recognition.
\newblock {\em IEEE Transactions on Affective Computing}, 13(3):1541--1554,
  2022.

\bibitem{saeed2020sense}
Aaqib Saeed, Victor Ungureanu, and Beat Gfeller.
\newblock Sense and learn: Self-supervision for omnipresent sensors.
\newblock {\em Machine Learning with Applications}, 6:100152, 2021.

\bibitem{aggarwal2019adversarial}
Karan Aggarwal, Shafiq Joty, Luis Fernandez-Luque, and Jaideep Srivastava.
\newblock Adversarial unsupervised representation learning for activity
  time-series.
\newblock In {\em Proceedings of the AAAI Conference on Artificial
  Intelligence}. AAAI Press, 2019.

\bibitem{banville2020uncovering}
Hubert~J. {Banville}, Omar {Chehab}, Aapo {Hyvärinen}, Denis-Alexander
  {Engemann}, and Alexandre {Gramfort}.
\newblock Uncovering the structure of clinical eeg signals with self-supervised
  learning.
\newblock {\em Journal of Neural Engineering}, 18(4):46020, 2021.

\bibitem{yue2022ts2vec}
Zhihan Yue, Yujing Wang, Juanyong Duan, Tianmeng Yang, Congrui Huang, Yunhai
  Tong, and Bixiong Xu.
\newblock Ts2vec: Towards universal representation of time series.
\newblock In {\em Proceedings of the AAAI Conference on Artificial
  Intelligence}, volume~36, pages 8980--8987, 2022.

\bibitem{icml2022iterative}
Ling Yang and Shenda Hong.
\newblock Unsupervised time-series representation learning with iterative
  bilinear temporal-spectral fusion.
\newblock In {\em Proceedings of the 39th International Conference on Machine
  Learning}, volume 162 of {\em Proceedings of Machine Learning Research},
  pages 25038--25054. PMLR, 17--23 Jul 2022.

\bibitem{zhang2022_tfConsistency}
Xiang Zhang, Ziyuan Zhao, Theodoros Tsiligkaridis, and Marinka Zitnik.
\newblock Self-supervised contrastive pre-training for time series via
  time-frequency consistency.
\newblock In {\em Advances in Neural Information Processing Systems}, 2022.

\bibitem{STFNets}
Dongxin Liu, Tianshi Wang, Shengzhong Liu, Ruijie Wang, Shuochao Yao, and Tarek
  Abdelzaher.
\newblock Contrastive self-supervised representation learning for sensing
  signals from the time-frequency perspective.
\newblock In {\em International Conference on Computer Communications and
  Networks (ICCCN)}, pages 1--10, 2021.

\bibitem{teblstm}
Qingchang Zhu, Zhenghua Chen, and Yeng~Chai Soh.
\newblock A novel semisupervised deep learning method for human activity
  recognition.
\newblock {\em IEEE Transactions on Industrial Informatics}, 15(7):3821--3830,
  2019.

\bibitem{gan_regression}
Mehdi Rezagholiradeh and Md~Akmal Haidar.
\newblock Reg-gan: Semi-supervised learning based on generative adversarial
  networks for regression.
\newblock In {\em IEEE International Conference on Acoustics, Speech and Signal
  Processing (ICASSP)}, pages 2806--2810, 2018.

\bibitem{success_semi}
Krist{\'o}f Marussy and Krisztian Buza.
\newblock Success: A new approach for semi-supervised classification of
  time-series.
\newblock In {\em Artificial Intelligence and Soft Computing}, pages 437--447,
  2013.

\bibitem{selfMatch}
Huanlai Xing, Zhiwen Xiao, Dawei Zhan, Shouxi Luo, Penglin Dai, and Ke~Li.
\newblock Selfmatch: Robust semisupervised time-series classification with
  self-distillation.
\newblock {\em International Journal of Intelligent Systems},
  37(11):8583--8610, 2022.

\bibitem{semiTime}
Haoyi Fan, Fengbin Zhang, Ruidong Wang, Xunhua Huang, and Zuoyong Li.
\newblock Semi-supervised time series classification by temporal relation
  prediction.
\newblock In {\em IEEE International Conference on Acoustics, Speech and Signal
  Processing (ICASSP)}, pages 3545--3549, 2021.

\bibitem{wang2017time}
Zhiguang Wang, Weizhong Yan, and Tim Oates.
\newblock Time series classification from scratch with deep neural networks: A
  strong baseline.
\newblock In {\em International Joint Conference on Neural Networks (IJCNN)},
  pages 1578--1585, 2017.

\bibitem{NIPS2017_3f5ee243}
Ashish Vaswani, Noam Shazeer, Niki Parmar, Jakob Uszkoreit, Llion Jones,
  Aidan~N Gomez, \L~ukasz Kaiser, and Illia Polosukhin.
\newblock Attention is all you need.
\newblock In I.~Guyon, U.~Von Luxburg, S.~Bengio, H.~Wallach, R.~Fergus,
  S.~Vishwanathan, and R.~Garnett, editors, {\em Advances in Neural Information
  Processing Systems}, volume~30. Curran Associates, Inc., 2017.

\bibitem{wang-etal-2019-learning}
Qiang Wang, Bei Li, Tong Xiao, Jingbo Zhu, Changliang Li, Derek~F. Wong, and
  Lidia~S. Chao.
\newblock Learning deep transformer models for machine translation.
\newblock In {\em Proceedings of the 57th Annual Meeting of the Association for
  Computational Linguistics}, pages 1810--1822, Florence, Italy, July 2019.
  Association for Computational Linguistics.

\bibitem{devlin2018bert}
Jacob Devlin, Ming-Wei Chang, Kenton Lee, and Kristina Toutanova.
\newblock {BERT}: Pre-training of deep bidirectional transformers for language
  understanding.
\newblock In {\em Proceedings of the Conference of the North {A}merican Chapter
  of the Association for Computational Linguistics: Human Language
  Technologies, Volume 1 (Long and Short Papers)}, pages 4171--4186,
  Minneapolis, Minnesota, June 2019. Association for Computational Linguistics.

\bibitem{khosla2020supervised}
Prannay Khosla, Piotr Teterwak, Chen Wang, Aaron Sarna, Yonglong Tian, Phillip
  Isola, Aaron Maschinot, Ce~Liu, and Dilip Krishnan.
\newblock Supervised contrastive learning.
\newblock In H.~Larochelle, M.~Ranzato, R.~Hadsell, M.F. Balcan, and H.~Lin,
  editors, {\em Advances in Neural Information Processing Systems}, volume~33,
  pages 18661--18673. Curran Associates, Inc., 2020.

\bibitem{anguita2013public}
Davide {Anguita}, Alessandro {Ghio}, Luca {Oneto}, Xavier {Parra}, and
  Jorge~Luis {Reyes-Ortiz}.
\newblock A public domain dataset for human activity recognition using
  smartphones.
\newblock In {\em European Symposium on Artificial Neural Networks}, pages
  437--442, 2013.

\bibitem{goldberger2000physiobank}
Ary~L. {Goldberger}, Luis A.~N. {Amaral}, Leon {Glass}, Jeffrey~M. {Hausdorff},
  Plamen~Ch. {Ivanov}, Roger~G. {Mark}, Joseph~E. {Mietus}, George~B. {Moody},
  Chung-Kang {Peng}, and H.~Eugene {Stanley}.
\newblock Physiobank, physiotoolkit, and physionet components of a new research
  resource for complex physiologic signals.
\newblock {\em Circulation}, 101(23):215--220, 2000.

\bibitem{supratak2017deepsleepnet}
Akara Supratak, Hao Dong, Chao Wu, and Yike Guo.
\newblock Deepsleepnet: A model for automatic sleep stage scoring based on raw
  single-channel eeg.
\newblock {\em IEEE Transactions on Neural Systems and Rehabilitation
  Engineering}, 25(11):1998--2008, 2017.

\bibitem{emadeldeen_attnSleep}
Emadeldeen Eldele, Zhenghua Chen, Chengyu Liu, Min Wu, Chee-Keong Kwoh, Xiaoli
  Li, and Cuntai Guan.
\newblock An attention-based deep learning approach for sleep stage
  classification with single-channel eeg.
\newblock {\em IEEE Transactions on Neural Systems and Rehabilitation
  Engineering}, 29:809--818, 2021.

\bibitem{PhysRevE.64.061907}
Ralph~G. Andrzejak, Klaus Lehnertz, Florian Mormann, Christoph Rieke, Peter
  David, and Christian~E. Elger.
\newblock Indications of nonlinear deterministic and finite-dimensional
  structures in time series of brain electrical activity: Dependence on
  recording region and brain state.
\newblock {\em Phys. Rev. E}, 64:061907, Nov 2001.

\bibitem{lessmeier2016condition}
Christian Lessmeier, James~Kuria Kimotho, Detmar Zimmer, and Walter Sextro.
\newblock Condition monitoring of bearing damage in electromechanical drive
  systems by using motor current signals of electric motors: A benchmark data
  set for data-driven classification.
\newblock In {\em European conference of the prognostics and health management
  society}, 2016.

\bibitem{ragab2020adversarial}
Mohamed Ragab, Zhenghua Chen, Min Wu, Haoliang Li, Chee-Keong Kwoh, Ruqiang
  Yan, and Xiaoli Li.
\newblock Adversarial multiple-target domain adaptation for fault
  classification.
\newblock {\em IEEE Transactions on Instrumentation and Measurement}, 70:1--11,
  2021.

\bibitem{mean_teacher}
Antti Tarvainen and Harri Valpola.
\newblock Mean teachers are better role models: Weight-averaged consistency
  targets improve semi-supervised deep learning results.
\newblock In {\em Proceedings of the 31st International Conference on Neural
  Information Processing Systems}, NIPS'17, page 1195–1204, Red Hook, NY,
  USA, 2017. Curran Associates Inc.

\bibitem{DivideMix}
Junnan Li, Richard Socher, and Steven~C.H. Hoi.
\newblock Dividemix: Learning with noisy labels as semi-supervised learning.
\newblock In {\em International Conference on Learning Representations}, 2020.

\bibitem{wen2021time}
Qingsong Wen, Liang Sun, Fan Yang, Xiaomin Song, Jingkun Gao, Xue Wang, and
  Huan Xu.
\newblock Time series data augmentation for deep learning: A survey.
\newblock In Zhi-Hua Zhou, editor, {\em Proceedings of the Thirtieth
  International Joint Conference on Artificial Intelligence, {IJCAI-21}}, pages
  4653--4660. International Joint Conferences on Artificial Intelligence
  Organization, 8 2021.
\newblock Survey Track.

\end{thebibliography}

\clearpage

\section*{Supplementary Material}
\renewcommand\thefigure{S.\arabic{figure}}
\renewcommand{\thesection}{S.\Roman{section}} 
\renewcommand\thetable{S.\arabic{table}}
\setcounter{figure}{0}
\setcounter{table}{0}
\setcounter{section}{0}

\section{Datasets Description}
We provide more details about the seven UCR repository datasets.

\subsection{Wafer}
Wafer dataset is a collection of several sensor measurements during the inline process control of silicon wafers for semiconductor fabrication. Each sample represents the recordings by one sensor during the processing of one wafer by one tool. The dataset contains two imbalanced classes, i.e., normal and abnormal, where only 10.7\% and 12.1\% of the train and test sets belong to the abnormal class.
    
\subsection{FordA and FordB}
These two datasets were used in a competition at the IEEE World Congress on Computational Intelligence, 2008. The timesteps in both datasets represent 500 measurements of engine noise, and the objective was to classify whether a certain symptom in an automotive subsystem exists or not.
For \textbf{FordA} dataset, the data was collected under minimal noise contamination, while \textbf{FordB} was collected in noisy conditions.

\subsection{PhalangesOutlinesCorrect (POC) and ProximalPhalanxOutlineCorrect (PPOC)}
Both datasets are extracted from a dataset designed to test the efficiency of the hand and bone outlines in bone age prediction. Some algorithms were applied to over 1,300 images to automatically extract hand outlines besides the outlines of three bones of the middle finger, i.e., proximal, middle, and distal phalanges. This generated three classification problems including \textbf{ProximalPhalanxOutlineCorrect}, while 
\textbf{PhalangesOutlinesCorrect} is the concatenation of these three problems.
The labels in both datasets indicate whether the image outlining is correct or incorrect. 

\subsection{StarLightCurves}
This dataset relates to an astronomical study for starlight curves, and the time-series represents the brightness of a celestial object as a function of time. Each curve represents an example from one of three classes: Classical Type-I Cepheids, Eclipsing Binaries (EB), RRab, and RRc RRLyrae (RRL).

\subsection{ElectricDevices}
This dataset was collected to study the usage behavior of consumers to the electricity in homes to reduce the carbon footprint in the UK. The readings were collected from 251 households.

\section{Experiments}
\subsection{Ensemble of Multiple Self-supervised Algorithms}
\begin{figure*}
    \centering
    \includegraphics[width=0.7\textwidth]{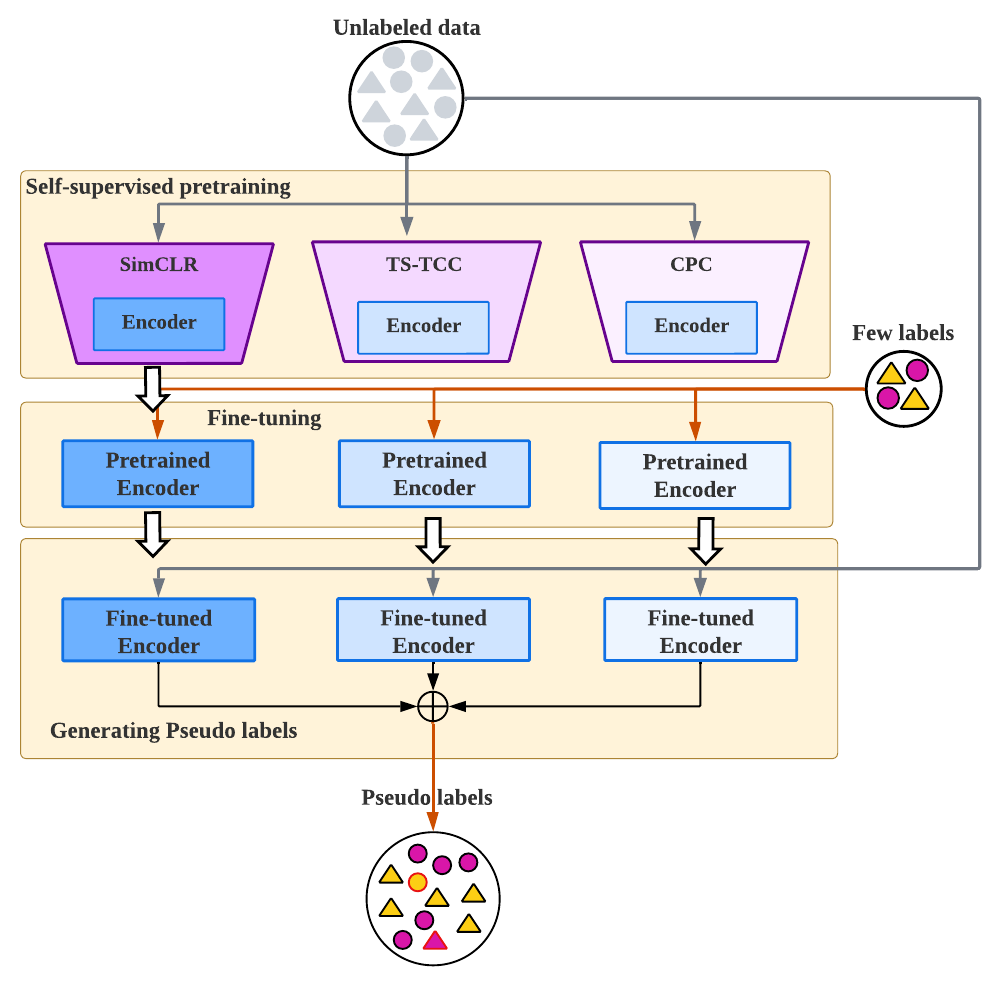}
    \caption{The process of pseudo labels generation from the ensemble of multiple self-supervised learning algorithms. We first pretrain each individual self-supervised learning algorithm with randomly-initialized encoders in Phase~1. Next, we fine-tune the pretrained encoders with the few labeled data in Phase~2. Following that, in Phase~3, we average the features of the three fine-tuned encoders to generate the pseudo labels to be used later in Phase~4.}
    \label{fig:ensemble_exp}
\end{figure*}

We conducted additional experiments by stacking the three best-performing self-supervised algorithms, i.e., SimCLR, CPC, and TS-TCC to produce pseudo labels for the semi-supervised training.
Fig.~\ref{fig:ensemble_exp} shows the process of training these algorithms and their ensemble to generate the pseudo labels. Specifically, we first pretrain each model with the unlabeled data, then fine-tune the pretrained encoders with the few labels. Subsequently, we average the features of the three fine-tuned encoders and use them to generate the pseudo labels. Last, the semi-supervised training in Phase~4 is performed with the fine-tuned TS-TCC encoder.

\begin{figure*}[b]
     \centering
     \begin{subfigure}[b]{0.47\textwidth}
         \centering
         \includegraphics[width=\textwidth]{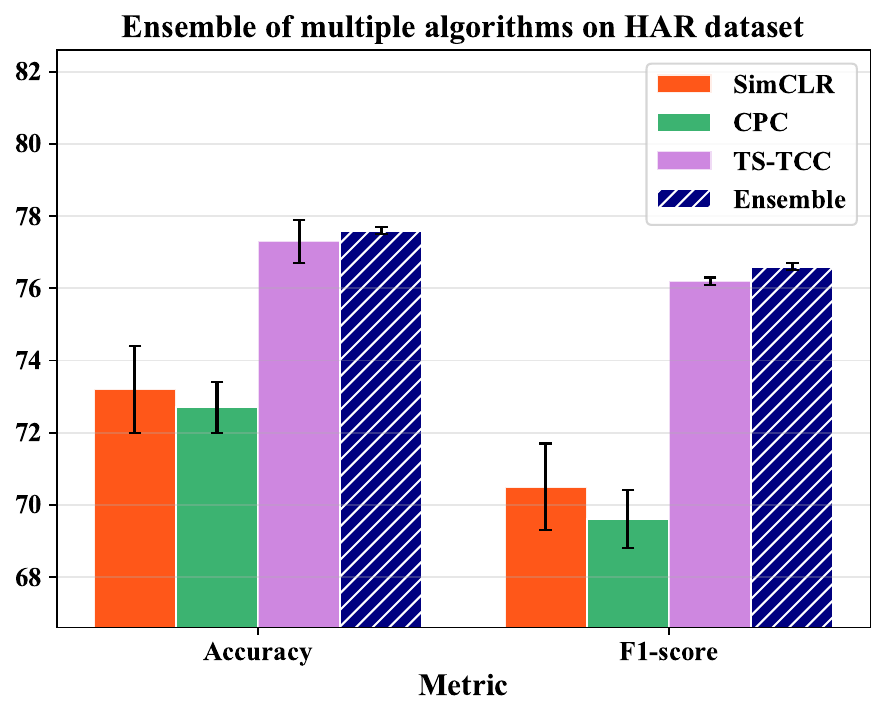}
         \caption{HAR dataset}
         \label{fig:ensemble_har}
     \end{subfigure}
     \hfill
     \begin{subfigure}[b]{0.47\textwidth}
         \centering
         \includegraphics[width=\textwidth]{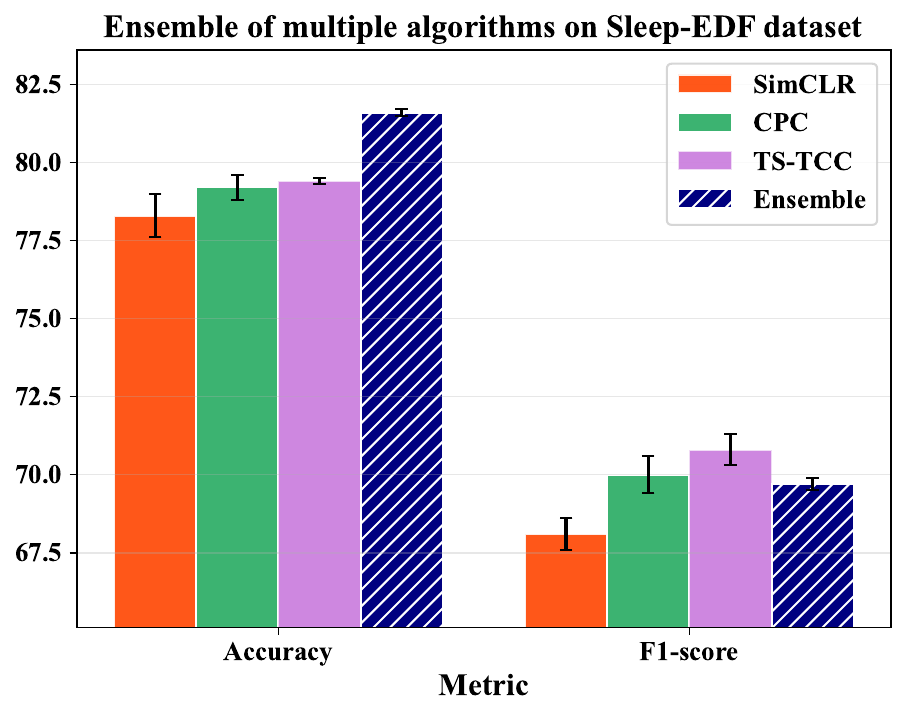}
         \caption{Sleep-EDF dataset}
         \label{fig:ensemble_sleep_edf}
     \end{subfigure}
\caption{Results of the ensemble of multiple self-supervised learning algorithms in Phase~1 against each individual algorithm. The ensemble showed a minor performance improvement.}
\label{Fig:ensemble}
\end{figure*}

The experimental results on HAR and Sleep-EDF datasets, shown in Fig.~\ref{Fig:ensemble}, demonstrate that the ensemble model has a superior performance and better stability, as indicated by the low standard deviation. We attribute this to the model's ability to generate high-quality pseudo labels consistently across different runs/seeds from the averaged features from the three models. However, the improvement is minor, and it comes at the cost of increased complexity, which can lead to longer training times and higher computational demands. As a potential extension of this work, we could explore ways to improve performance while controlling for training time and complexity.

\end{document}